\documentclass[pmlr,twocolumn,10pt]{jmlr} 

\usepackage{longtable}
\usepackage{multirow}

\usepackage{booktabs}
\usepackage{siunitx}
\usepackage{hyperref}
\hypersetup{
    colorlinks=true,
    linkcolor=blue,
    filecolor=magenta,      
    urlcolor=cyan,
    }

\newcommand{\equal}[1]{{\hypersetup{linkcolor=black}\thanks{#1}}}

\theorembodyfont{\upshape}
\theoremheaderfont{\scshape}
\theorempostheader{:}
\theoremsep{\newline}

\jmlrvolume{LEAVE UNSET}
\jmlryear{2022}
\jmlrsubmitted{LEAVE UNSET}
\jmlrpublished{LEAVE UNSET}
\jmlrworkshop{Conference on Health, Inference, and Learning (CHIL) 2022} 

\title[Real-Time Seizure Detection using EEG]{Real-Time Seizure Detection using EEG: \\A Comprehensive Comparison of Recent Approaches \\under a Realistic Setting}

\author{%
\Name{Kwanhyung Lee}\equal{These authors contributed equally} \Email{kwanlee9209@aitrics.com}\\
\addr AITRICS, Republic of Korea
\AND
\Name{Hyewon Jeong}\footnotemark[1]\nametag{\thanks{This work was done while the authors were in AITRICS}} \Email{hyewonj@mit.edu}\\
\addr Massachusetts Institute of Technology, USA \\
\addr AITRICS, Republic of Korea
\AND
\Name{Seyun Kim}\footnotemark[2] \Email{kim79@cooper.edu}\\
\addr The Cooper Union for the Advancement of Science and Art, USA \\ 
\addr AITRICS, Republic of Korea
\AND
\Name{Donghwa Yang} \Email{soul2star@nhimc.or.kr}\\
\addr National Health Insurance Service Ilsan Hospital, Republic of Korea \\
Severance Children’s Hospital, Yonsei University College of Medicine, Republic of Korea
\AND
\Name{Hoon-Chul Kang} \Email{hipo0207@yuhs.ac}\\
\addr Severance Children’s Hospital, Yonsei University College of Medicine, Republic of Korea
\AND
\Name{Edward Choi} \Email{edwardchoi@kaist.ac.kr}\\
\addr KAIST, Republic of Korea
}

\begin{document}

\maketitle
\begin{abstract}
    Electroencephalogram (EEG) is an important diagnostic test that physicians use to record brain activity and detect seizures by monitoring the signals. There have been several attempts to detect seizures and abnormalities in EEG signals with modern deep learning models to reduce the clinical burden. However, they cannot be fairly compared against each other as they were tested in distinct experimental settings. Also, some of them are not trained in real-time seizure detection tasks, making it hard for on-device applications.
    In this work, for the first time, we extensively compare multiple state-of-the-art models and signal feature extractors in a real-time seizure detection framework suitable for real-world application, using various evaluation metrics including a new one we propose to evaluate more practical aspects of seizure detection models.
\end{abstract}
\section{Dataset and Code Availability}
The data used for our work can be found from the Temple University Hospital (TUH) EEG Seizure Corpus (TUSZ) dataset V1.5.2 (\url{https://isip.piconepress.com/projects/tuh_eeg/html/downloads.shtml}). Our code is available at \url{https://github.com/AITRICS/EEG_real_time_seizure_detection}.
\section{Introduction}
Epilepsy is a neurologic disorder characterized by epileptic seizures \citep{fisher2014ilae} and the precise frequency and type of seizures of the patient should be identified in order to diagnose epilepsy. This affects the decisions of pharmacological treatment, dietary therapy, and surgical treatment \citep{scheffer2017ilae}. In this process, electroencephalogram (EEG) is an essential diagnostic tool for seizure detection. Inpatient video-EEG recording is a routine diagnostic method for capturing seizure events for deciding proper future treatment. However, many of the event onsets are prone to be missed on-site as clinicians cannot always be present by the patient, and they cannot rewind the whole 24-hour video to find out the actual onset \citep{nordli2006usefulness}.
Family members of a patient can press a button to denote the onset of the seizure event, but non-experts (\textit{i.e.} family member) can make false positive or false negative decision.

In order to accurately detect seizures for prompt intervention, we need an on-site seizure detector that can be used while EEG is being recorded.
This task can, in theory, be done by a well-trained neurologist where they routinely scan the recorded EEG signals to find out the characteristic signal for the potential diagnoses. 
This manual process, however, can take up to several hours, making it labor-intensive as well as untimely.
To improve the timeliness of the seizure detection and reduce the burden of clinicians, we need a machine that accurately detects the seizure signal in real-time to help physicians diagnose and prescribe.

Due to the need of an accurate seizure detector, several hospital-based research organizations released public benchmark datasets \citep{obeid2016temple, shah2018temple, shoeb2009application, goldberger2000physiobank, andrzejak2001indications} for the seizure detection task.
There have been several attempts to detect seizure and abnormalities with these public EEG signal datasets using deep neural network models \citep{acharya2018deep, roy2019chrononet, mohsenvand2020contrastive, golmohammadi2017gated, akbarian2020framework, zabihi2015analysis}.
More recent works have focused on multi-class seizure detection \citep{sriraam2019convolutional, daoud2019efficient, ahmedt2020neural, priyasad2021interpretable, jia2022variable}, and real-time seizure detection \citep{shawki2020deep, bomela2020real, thyagachandran2020seizure, fan2018detecting}. Although previous studies have explored multiple model architectures for seizure detection, we cannot fairly compare each of the model as they used distinct experimental settings and evaluation metrics.
Also, not all the studies adopted the real-time seizure detection tasks, limiting the practicality of the models.
 
\begin{figure*}[ht!]
	\centering
	\includegraphics[width=\textwidth]{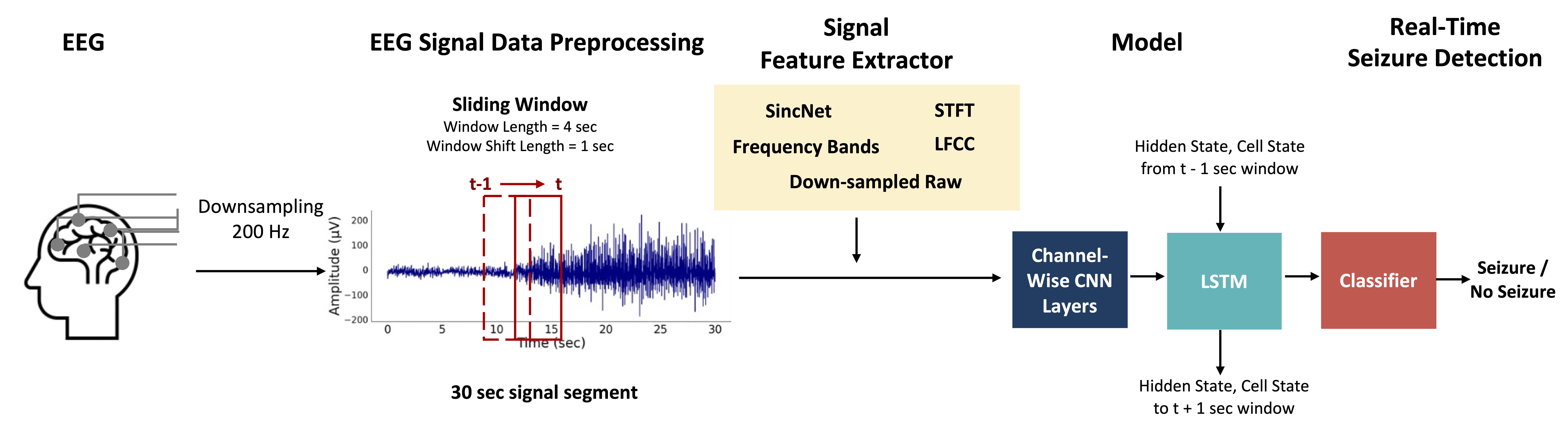}
    \caption{\small \textbf{Real-Time Seizure Detection with CNN (+LSTM) models.} We downsample the EEG signal and extract features (Figure \ref{preproc}). The models detect whether ictal/non-ictal signal appears within the 4-second sliding window input. We present an example case with a Raw EEG signal but other signal feature extractors can also be applied to the pipeline.}
	\label{concept}
\end{figure*}
In this paper, for the first time, we present and compare the performance of various state-of-the-art models and feature extraction strategies under the same experimental setting of real-time binary seizure detection (Figure \ref{concept}).
We performed the seizure (ictal) onset detection task, where the model receives EEG signal and detects whether the patient is currently in ictal phase or not, given that the entire process from feature extraction to detection must be completed within the stride time of the sliding window in order to guarantee real-time process.
We also proposed margin-based evaluation metric, which is an evaluation metric providing more useful information to physicians by assessing whether a model detects seizure events within the margin from the seizure onset and offset. 
Our work thus presents a comprehensive summary of various models' performance under a strictly practical and realistic setting, such that researchers and practitioners can directly refer to our study for deploying their own models in practice, or use our study as a benchmark for developing a new seizure detection model in the future.


     

\section{Background and Related Work}
\subsection{Signal Feature Extractor}
Signal processing and frequency extraction are useful tools for extracting features from raw signal, which has been used for preprocessing speech signal or biomedical signals (Electrocardiogram (ECG), Heart sound, or EEG signal). One of the popular feature extractors for medical signals is Short Time Fourier Transform (STFT) \citep{sejdic2009time, springer2015logistic, sriraam2019convolutional}, which applies a sequence of Fast Fourier transforms (FFT) to signals using a sliding window. Some of the previous works used specific frequency bands out of extracted important frequency features (e.g., delta wave (0-4 Hz), theta wave (4-8 Hz), or alpha wave (8-12 Hz)) that are related to brain activity \citep{parvez2014eeg, liu2019epileptic}. Other seizure detection models \citep{shawki2020deep, thyagachandran2020seizure, golmohammadi2019automatic} used linear scale filter bank on STFT output matrix, which is called Linear frequency cepstral coefficients (LFCC). A recent work \citep{priyasad2021interpretable} utilized SincNet \citep{ravanelli2018speaker} filters onto channel-wise one dimensional sigal input for seizure detection task. SincNet extracts signal features with Convolutional Neural Networks (CNN) Sinc function filters, without traditional feature extracion methods. As recent deep learning architectures such as CNN provide flexible and powerful feature extraction, recent works presented high performing deep learning models using raw signal without feature extraction in various tasks including speech recognition (Wav2vec2.0 \citep{baevski2020wav2vec}) and EEG seizure detection task \citep{roy2019chrononet, mohsenvand2020contrastive}). \cite{mohsenvand2020contrastive} downsampled the raw input signal to multiple frequencies to extract more diverse feature information. We extracted signal features from EEG input with these diverse signal feature extraction methods (Figure \ref{preproc}) that were frequently used for biomedical signal or speech processing. 

\subsection{Seizure Detection}
Recent studies introduced deep neural network models to binary and multi-class EEG seizure detection and classification. In this paper, we are focusing on binary seizure detection for simpler setting. A simple seizure detector with CNN was proposed to detect seizure \citep{andrzejak2001indications, acharya2018deep, yuan2018novel}. Other works used Long Short Term Memory (LSTM) \cite{hochreiter1997long} based architectures to capture temporal dependencies. \cite{golmohammadi2017gated} compared Gated Recurrent Unit (GRU) and LSTM for real-time binary seizure detection. Another work encoded the graph theoretic relationship between EEG leads into features to detect seizure onset time \citep{akbarian2020framework, zabihi2015analysis} on CHB-MIT dataset \citep{shoeb2009application, goldberger2000physiobank}. Extending binary detection models, there has been recent attempts to classify multiple seizure types using CNN architecture \citep{asif2020seizurenet, sriraam2019convolutional, daoud2019efficient, ahmedt2020neural, priyasad2021interpretable, jia2022variable}, including AlexNet \citep{krizhevsky2012imagenet, sriraam2019convolutional}, ResNet \citep{he2016deep}, DenseNet \citep{huang2017densely}, and MobileNetV3 \citep{howard2019searching}. Some of the works detected abnormal signals in input EEG signal \citep{roy2019chrononet, mohsenvand2020contrastive}. Rather than classifying the signal segment to single or multiple seizure types, \cite{daoud2019efficient} predicted seizure events before it occurs. However, many of the previous models cannot be applied for real-time seizure detection because they are trained to detect seizure events on the whole EEG signal. 

\subsection{Real-Time Seizure Detector}
More recent works focused on building real-time seizure detectors, using various experimental settings. Channel-wise LSTM was used in combination with LFCC as feature extractor \citep{shawki2020deep}, and Time Delay Neural Networks (TDNN) with LSTM using LFCC \citep{thyagachandran2020seizure} was applied on Temple University Hospital (TUH) dataset \citep{obeid2016temple}, one of the public benchmark EEG dataset for seizure detection task. Some of the real-time seizure detectors are inappropriate for clinical use as they are slow detector. \cite{shawki2020deep} achieved computation speed of 1.81 seconds per 1 second sliding window with CPU processor, which is too slow for real-time processing. Another work \citep{bomela2020real} did not report the latency of a detector. Also some used small-scale CHB-MIT dataset that includes EEG signal of only 25 patients \citep{fan2018detecting, wang2018hardware}.

\subsection{Evaluation Metric for Seizure Detection}
The performance of a model on EEG seizure detection can be measured by evaluating how well a model detected seizure events compared to the event label. Therefore, the evaluation metric should be valid for the seizure detection task \citep{ziyabari2017objective}, distinct from the accuracy measure used in time-series data or the word error rate used in the speech recognition task \citep{wang2003word, mostefa2006evaluation}. Scoring the performance of a model based on the event is called term-based scoring, and metrics evaluating per channel and per unit window is called the epoch-based method. Any-Overlap (OVLP) \citep{wilson2003seizure, gotman1997automatic} is a term-based scoring method that measures True Positive (TP) by counting the overlapping seizure detection hypothesis and label. OVLP is thus more permissive, outputs high sensitivity as it only considers the event-based scoring, not considering the duration of the event. Time-Aligned Event Scoring (TAES) improves this method by weighting the amount of overlap between label and model prediction. These two methods assess the model prediction within the whole signal, which is not suitable for real-time seizure detection task. Epoch-Based Sampling (EPOCH) \citep{baldassano2016novel} provides a more relevant metric for the real-time seizure detection, which measures the overlap between label and prediction within a unit time window (that is a data point within an epoch), per EEG channel. In this work, we propose a new metric called MARGIN which evaluates the detection performance of the seizure detection model within onset/offset margin. Other than these four evaluation metric, we additionally measured average onset latency time and calculated the average onset latency time of seizure events. Summary of each metric is in Figure \ref{fig_eval}.
\section{Methods}
\label{sec:methods}
\begin{figure*}[h!]
	\centering
	\includegraphics[width=\textwidth]{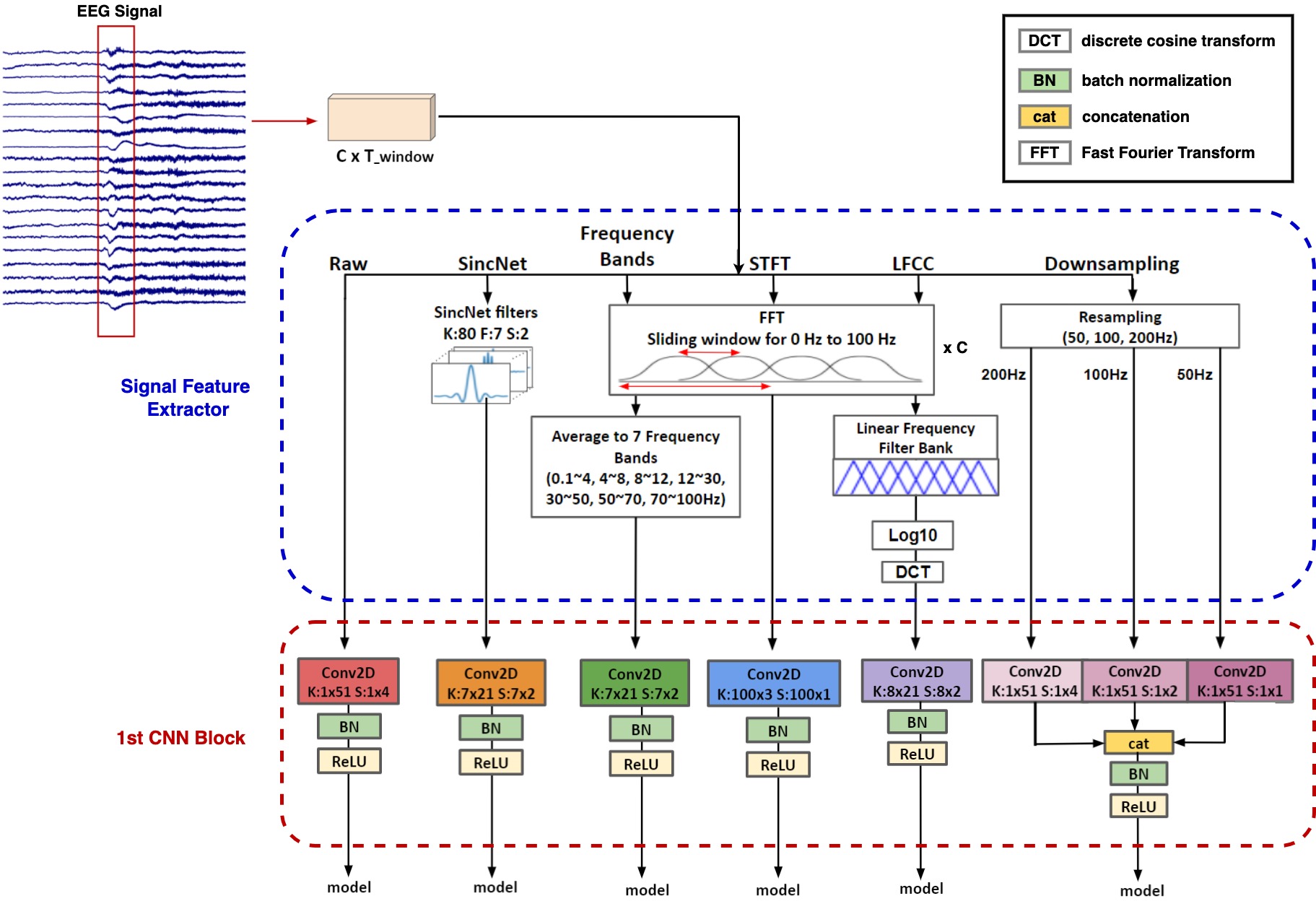}
    \caption{\small \textbf{Signal Feature Extractors} and input preprocessing. Upon downsampling EEG signal with 200Hz, the input undergoes one of the signal feature extractors and fed into the first CNN2D layer designed for each extractors. C: number of EEG signal channels, T\_window: length of one sliding window, K: Kernel size, F: Frequency bands, S: Stride}
	\label{preproc}
	\vspace{-5pt}
\end{figure*}
\subsection{Dataset}
We used the public benchmark dataset, the Temple University Hospital EEG Seizure Data Corpus (V1.5.2) \citep{obeid2016temple} (TUH v 1.5.2), which contains the largest number of seizure types and patient instances. We provide better experimental setting with the largest dataset as some of the prior works used private datasets \citep{daoud2019efficient} and small-scale public datasets \citep{akbarian2020framework, fan2018detecting, zabihi2015analysis} which makes it difficult to reproduce and generalize the results. 

\begin{table}[ht]
	\small
	\centering
	\caption{\textbf{Dataset Organization.} The number of signal types in the whole dataset after slicing the whole signal segment into 30 second signal segments.}
	\label{dataset}
	\begin{tabular}{c|c}
		\toprule
		Signal Types & Sample No.\\
		&(Train/Val/Test)\\
		\midrule
		Non-ictal (Patients) & 44,599 / 9,221 / 6,907\\
		Non-ictal (Normal Control) & 37,017 / 1,051 / 822\\
		Ictal, Non-ictal Signal & 3,636  / 636 / 541\\
		Ictal & 3,671 / 964 / 456\\
		\bottomrule
	\end{tabular}
\end{table}
The number of EEG signal types and patient number are summarized in Table \ref{dataset} and Appendix Table \ref{tuh}. Among total 7,034 signal streaks from 642 patients, there were total 304 patients with 3,047 signal streaks which each includes one or more seizure episode (Patients group in Table \ref{dataset}). The dataset also included 338 patients in Normal Control group, which is the group without any seizure events. There were total 1,838 normal signals from Normal Control, and 2,149 normal signals from Patients group. We have 589 subjects in the training dataset and 51 subjects in the development dataset (640 subjects), which we divide into validation (25 subjects) and test set (26 subjects) according to patient instances.
In order to prevent dataset shift, we randomly sample signal types within a batch so that each batch contains equal number of signal type (Table \ref{dataset}, Appendix Figure \ref{sampling}). Additional details of the dataset, including slight modification from the original TUH 1.5.2 open data, can be found in Appendix Section \ref{sec: app_dataset}.

The original EEG signal data was uniformly re-sampled with the sampling rate of $200$ Hz to standardize the sampling rate of all signals (Section \ref{sec:method_featext} 6) for more detail). We sampled the EEG signals from the leads listed in Figure \ref{leadinfo} for different types of EEG montage (Unipolar, Transverse Bipolar Montage). We simply subtracted the voltage between Unipolar leads to acquire Bipolar montage signal data, following Bipolar lead info from SeizureNet \citep{asif2020seizurenet} (Appendix Figure \ref{leadinfo}, Table (a)).

\subsection{Real-Time Seizure Detection}

The real-time seizure detector explained in Figure \ref{concept} can be directly used for clinical setting, where the model receives signal input of 4 seconds window on-device every 1 second (\textit{i.e.} window shift length). Thus, we would want a model to process each window within the shift length (1 second). We prepare each downsampled dataset into the size of window length (4 seconds), and slided the window with 1 second window shift length. We found the optimal window length by considering processing speed, and optimal shift length by considering the performance (Appendix Section \ref{sec:windowshift}).

After preparing EEG input, we extracted features with signal feature extractors (Section \ref{sec:method_featext}, Table \ref{feature}), and fed the extracted signal into each model (Section \ref{sec:method_model}) to obtain the detected output. 
The models were mainly evaluated with EPOCH metric and MARGIN as it provides per-window evaluation (EPOCH) and measures practicality of the model as real-time seizure detector (MARGIN); we also explored other evaluation metrics on all 15 models (Appendix Table \ref{eval_models}).

\subsection{Signal Feature Extractor}
\label{sec:method_featext}
We explore six different signal feature extraction techniques to generate input data for CNN and LSTM combined models (Figure \ref{preproc}). All six methods affect the final performance of a model, and each has different processing speed and feature representation (Table \ref{feature}).
Note that different types of first CNN2D block was applied to the output of each feature extractor, which is described in detail in each section below. For more detailed information on signal feature extractor (Section \ref{sec: featextractor}), CNN1D input (Figure \ref{1D_architecture}), experimental result with various models (Table \ref{eval_models}), please see Appendix.

\noindent\textbf{1) Raw}: Downsampled raw EEG signal (without feature extraction) passed to a CNN2D block with kernel shape $1 \times 51$ with stride $1 \times 4$.

\noindent\textbf{2) SincNet} \citep{ravanelli2018speaker}: We performed time-domain convolution between the input EEG signal and SincNet filters in the first CNN2D layer to preserve more information from the input signal frequency. We used $7$ distinct SincNet bands with trainable frequency band range with the stride of $2$. CNN2D block with kernel shape $7 \times 21$ with stride of $7 \times 2$ was applied to channel-wise 2D input.

\noindent\textbf{3) Short Time Fourier Transform (STFT)}: Fast Fourier Transform (FFT) was applied with sliding window of frame length 0.125 seconds and frame shift length (hop length or overlapping size) across 50\% of frame length, following the Hann window function \citep{oppenheim1999discrete}. The extracted feature represents information from both frequency and temporal domain. STFT can be also further processed with various steps, such as Frequency Bands and LFCC.

\noindent\textbf{4) Frequency Bands}: STFT was applied to extract frequency time features over $7$ frequency bands: 1–4, 4–8, 8–12, 12–30, 30–50, 50–70, 70–100 Hz.
Features on each band is then averaged over a signal.
We follow the parameters from \cite{liu2019epileptic} and \cite{mirowski2009classification}, which used predefined frequency ranges. We applied the same CNN2D block from SincNet: kernel shape $7 \times 21$ with stride of $7 \times 2$. 

\noindent\textbf{5) Linear frequency cepstral coefficients (LFCC)}: We used 0.3 seconds window length and 0.15 seconds window shift length to generate 8 absolute LFCC features out of the feature extractor, following setup from \cite{shawki2020deep}.

\noindent\textbf{6) Downsampled Raw (50, 100, 200Hz)} We re-sampled the original 200Hz raw signal data to 100Hz and 50Hz raw signals. The three differently sampled signals (50, 100, and 200 Hz) are then fed to CNN block separately and concatenated for further training. Sampling rates of 50, 100, 200Hz are enough to represent the features as it is more than twice the characteristic frequency range of seizure signals (0.5-25 Hz) \citep{shoeb2009application}. Furthermore, anti-aliased and downsampled signal does not improve the performance of our model (Table \ref{antialiasing}). 


\subsection{Models}
\label{sec:method_model}
\begin{figure*}[h!]
	\centering
	\includegraphics[width=0.8\textwidth]{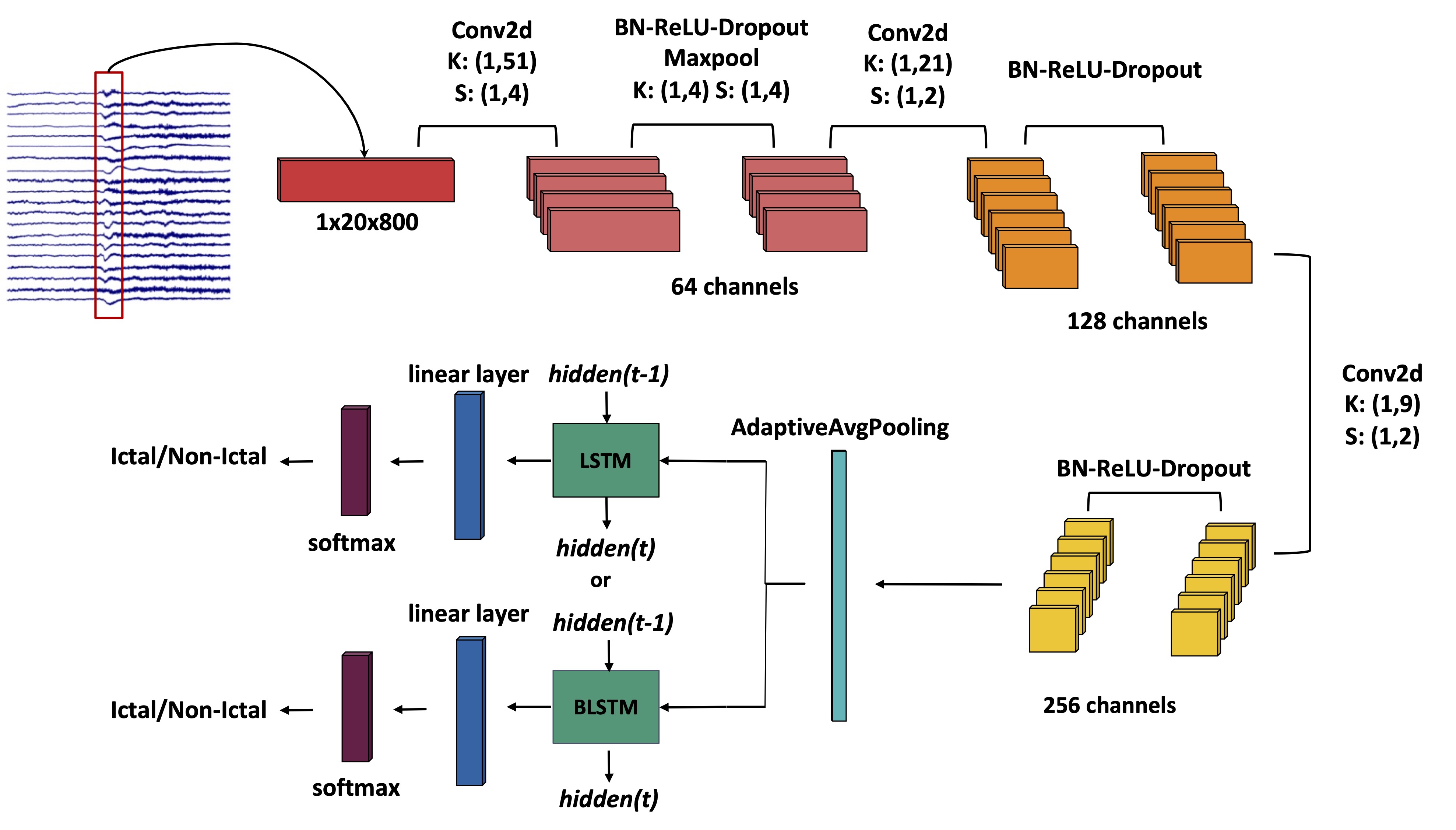}
    \caption{\small \textbf{Architecture of CNN2D + LSTM seizure detector} with raw EEG input. K: Kernel size, S: Stride}
	\label{architecture}
\end{figure*}
We extensively searched over $15$ architectures for real-time seizure onset detection on both Unipolar (Appendix Table \ref{unipolar}) and Bipolar (Table \ref{Result}) montage. We also measured the model size (Appendix Table \ref{modelsize}) and inference speed on CPU and GPU to analyze the feasibility of each model as the real time detector (Appendix section \ref{sec: speed}, Table \ref{Result}).
Details on the parameters and architectures of each model are summarized in Appendix Section \ref{sec: app_arch}, \ref{sec: size}, and Figure \ref{architecture}.

\subsubsection{CNN2D with LSTM}
\label{sec:CNN2DLSTM}
CNN2D with LSTM consists of three parts: CNN encoder for feature extraction, Recurrent Neural Network (RNN) module for temporal feature representation, and classifier (Figure \ref{concept}). After signal feature extraction, we applied CNN2D layer with 1D kernel (filters $\times$ number of channels $\times$ timestep) so that this shape filter is applied to each EEG channel separately, such that we can preserve channel-wise information within EEG signal. This also brings a model the faster feature extraction compared to the computation-heavier model with multiple channel-wise 1D CNN kernels \citep{priyasad2021interpretable, mohsenvand2020contrastive}. We tested the following five different CNN2D + LSTM models where CNN2D layer is followed by 2D adaptive average pooling module with RNN models, and then classifier block (Figure \ref{architecture}).
 
\noindent\textbf{1) CNN2D+LSTM}: As mentioned above, channel-wise CNN2D layer and LSTM are used for both channel-wise signal shape feature and temporal feature for CNN2D+LSTM model.\\
\noindent\textbf{2) CNN2D+BLSTM}: Same model architecture with 1) but alternatively with Bidirectional Long Short-Term Memory (BLSTM) in RNN module for temporal feature representation.\\
\noindent\textbf{3) ResNet-short+LSTM}: CNN2D+LSTM model where CNN blocks are from ResNet model \citep{he2016deep} helping a model learn both simple and rather complex shape for each seizure type with flexible gradient flow through skip connections.\\
\noindent\textbf{4) ResNet-short+Dilation+LSTM}: Almost same model architecture with 3) but three parallel Dilated Convolution (dilation of $1 \times 1$, $1 \times 2$, $1 \times 4$) was used in the first CNN2D block. The three outputs of the Dilated Convolution blocks are then concatenated and forwarded to CNN blocks from ResNet model as in 3). We expect the three dilated convolution would enable CNN block to receive diverse range of signals, which is similar to down-sampled raw \citep{mohsenvand2020contrastive} but with higher temporal resolution as the filter is shared across temporal axis. The more dilation the kernel has, the wider range of lower frequency signals the CNN model receives.\\
\noindent\textbf{5) MobileNetV3-short+LSTM}: We used MobileNetV3 \cite{howard2019searching} which is designed to perform on par with other CNN models but with small size, available for the off-line portable device. In this model, MobileNetV3 CNN blocks are used for feature extraction and connected to LSTM layers. 

\subsubsection{CNN1D with LSTM}
CNN1D with LSTM models follow the same structure as CNN2D + LSTM models (CNN, RNN, followed by Classifier layer) but with CNN1D block. Unlike previous EEG seizure type classification models \citep{priyasad2021interpretable, mohsenvand2020contrastive} which used multiple channel-wise CNN1D modules, our model does not have channel-wise CNN encoder. Our model rather has a CNN1D encoder with the filter size following the number of channels $\times$ timestep in order to increase the speed for real-time seizure detection.

\noindent\textbf{6) CNN1D+LSTM}: This model consists of a CNN 1D feature extraction block, followed by 1D adaptive average pooling module, LSTM layers, and classifier block (Figure \ref{1D_architecture}).\\
\textbf{7) CNN1D+BLSTM}: Same model architecture with 6) but with BLSTM instead of LSTM layer.

\subsubsection{Benchmark Models}
We adopted the original architecture of six different benchmark models frequently used for EEG seizure detection, abnormality detection, and seizure classification \citep{jia2022variable, ahmedt2020neural, seizuretype2020tuh, priyasad2021interpretable, asif2020seizurenet, sriraam2019convolutional, roy2019chrononet, thyagachandran2020seizure}.\\ 
\textbf{8) ResNet18} \citep{he2016deep}: We used ResNet18 among all ResNet variations for its small model size suitable for real time detection. \\
\textbf{9) MobileNetV3} \citep{howard2019searching}: CNN model with squeeze-and-excite block after expansion layer, designed to perform well on mobile devices. \\
\textbf{10) AlexNet} \citep{krizhevsky2012imagenet}: One of the early CNN architecture revolutionized CNN by using ReLU, dropout, overlapping pooling, and data augmentation which are now easily found in many CNN architectures. \\
\textbf{11) DenseNet} \citep{huang2017densely}: CNN model that connects feature map of every layer with the ones with following layers, and concatenate them.\\
\textbf{12) ChronoNet} \citep{roy2019chrononet}: This model uses a CNN1D layer parallelized with various kernels, which is then connected to GRU.\\
\noindent\textbf{13) TDNN with LSTM} \citep{thyagachandran2020seizure}: A combination of Time Delay Neural Networks (TDNN) and LSTM. TDNN layer performs temporal convolution by convolving each channel-wise EEG signal segment across timestep; LSTM layer then combines the channel-wise contextual representation to model sequence information. TDNN is commonly used for signal detection (EEG person identification \citep{kumar2019subspace}, voice activity detection, anomaly heart sound detection).

\subsubsection{Feature Transformers}
\noindent\textbf{14) Feature Transformer}: The model captures similarity between signals via self-attention on the channel-wise CNN features (Appendix Figure \ref{featatt}). Unlike previous works using attention module \citep{priyasad2021interpretable, yuan2018novel, yuan2018multi} that calculate importance of each EEG channel, we can extract the relationship between the channels from the measured channel-to-channel similiarity.

\begin{figure*}[h!]
	\centering
	\includegraphics[width=\textwidth]{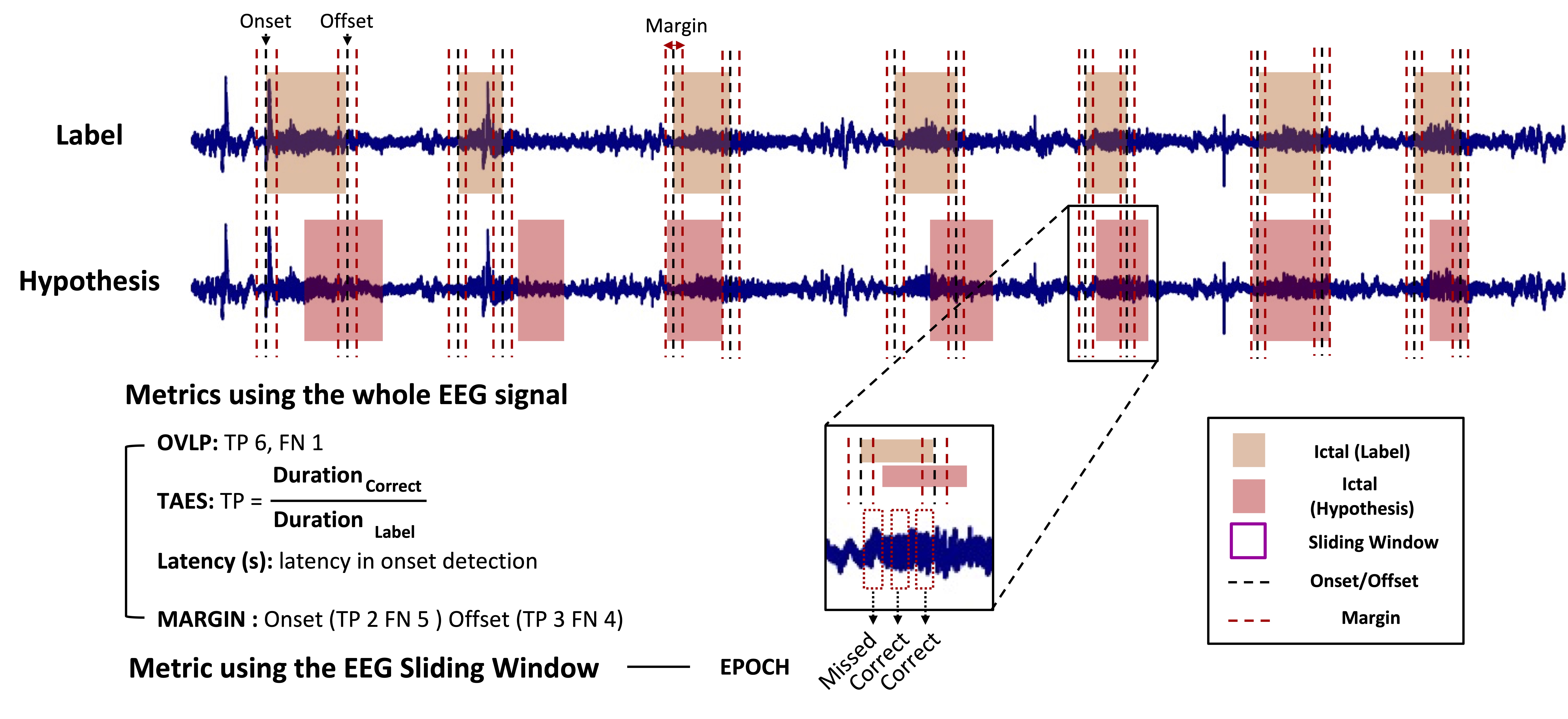}
    \caption{\small \textbf{Evaluation Methods} on example raw EEG signal. OVLP, TAES, MARGIN evaluates the model on the whole raw signal. OVLP outputs whether the hypothesis overlaps with the label, and TAES outputs the duration of how much hypothesis overlaps with label. MARGIN evaluates whether the onset/offset of event hypothesis is within set margin window. EPOCH outputs evaluation on hypothesis based on each window. Then final TP is measured by counting the number of windows where the model hypothesis was correct, divided by the total number of windows within each batch. TP - True Positive; FN - False Negative.}
	\label{fig_eval}
	\vspace{-10pt}
\end{figure*}
\noindent\textbf{15) Guided Feature Transformer}: The model considers the neighboring connection between all pairs of EEG leads by simply multiplying the channel adjacency mask (channel relation mask) to the self attention map (Appendix Figure \ref{featatt}).
Examples of adjacent EEG leads with C3 are summarized in Appendix Figure \ref{leadinfo}, where we denote neighboring leads to be connected in Unipolar montage and directly used the connection between leads in Bipolar montage.

\subsection{Evaluation Metrics}
\label{sec: evalmetric}
We calculate the binary classification performance (\textit{e.g.} AUROC, AUPRC) of the models on the test dataset, using four different methods to determine correct and incorrect classifications (Figure \ref{fig_eval}).
To evaluate sensitivity (True Positive Rate, TPR) and specificity (True Negative Rate, TNR) using OVLP, TAES, EPOCH and onset/offset accuracy of MARGIN, we measured values over the whole EEG signal data of a patient where the model is trained with sliding window. In addition to four evaluation metrics introduced below, we measured the onset latency, which is the average latency from the label onset to hypothesis onset. 


\noindent\textbf{1) Any-Overlap (OVLP)}: OVLP provides sensitivity permissive metric which measures whether the prediction overlaps with the label.

\noindent\textbf{2) Time-Aligned Event Scoring (TAES)}: TAES measures the amount of overlap between seizure label and prediction, by calculating the portion of correctly predicted among total duration of seizure label.

\noindent\textbf{3) Epoch-Based Sampling (EPOCH)}: EPOCH outputs a summary score per each unit window, which evaluates the accuracy of prediction by comparing with the ground truth label. Each epoch contains the preprocessed EEG signals within a fixed time window.

\noindent\textbf{4) Margin-Based Boundary Evaluation (MARGIN)}: We propose a new evaluation metric for EEG seizure detection task. MARGIN evaluates whether a model has detected the seizure within a fixed margin (3 seconds, 5 seconds) before and after the onset / offset time of seizure. Thus, a model with higher MARGIN accuracy would more accurately detect the onset and offset of seizure event, thus would help prompt clinical intervention. MARGIN also compensates for the inherent measurement error in scalp EEG signal, as the signal itself is an indirect measure of neural activity. Because of this uncertainty in the measured signal, actual seizure can often be initiated earlier than the onset diagnosed by physicians. We thus give the model a safety margin to take into account situation where a model detects a seizure event before the onset.
\section{Results}
\label{sec:results}
\begin{table*}[ht!]
	\tiny
	\centering
	\caption{\textbf{Result} of real-time seizure detection on \textit{raw} bipolar TUH EEG dataset trained with each architecture, averaged over $5$ runs. Please see Appendix section \ref{sec: speed} for CPU and GPU settings. We report TPR, TNR, and measured MARGIN when TNR is above 0.95.}
	\label{Result}
	\begin{tabular}{c|cccccc}
		\toprule
		Methods & AUROC & AUPRC & TPR & TNR & MARGIN(5sec)  &  CPU/GPU\\
		&&&&& Acc(Onset/Offset) & Speed (sec)\\
		\midrule
		CNN2D + LSTM &  0.89 $\pm$ 0.01  &  0.88 $\pm$ 0.01 &  0.81 $\pm$ 0.03  &  0.83 $\pm$ 0.03 & 0.56 / 0.51  & \textbf{0.079 / 0.004} \\
		CNN2D + BLSTM &  0.87 $\pm$ 0.01  &  0.86 $\pm$ 0.02  &  0.76 $\pm$ 0.05  &  0.85 $\pm$ 0.04 & 0.57 / 0.64  & 0.224 / 0.005 \\
	    ResNet-short + LSTM &  \textbf{0.92 $\pm$ 0.00}  &  \textbf{0.91 $\pm$ 0.00}  &  \textbf{0.83 $\pm$ 0.02}  &  \textbf{0.85 $\pm$ 0.01} & \textbf{0.62 / 0.65}  & 0.941 / 0.013 \\
		ResNet-short+Dilation + LSTM &  \textbf{0.91 $\pm$ 0.01} &  \textbf{0.90 $\pm$ 0.01} &  \textbf{0.84 $\pm$ 0.01} & \textbf{0.83 $\pm$ 0.03} & \textbf{0.58 / 0.61} & 0.682 / 0.031 \\
		MobileNetV3-short + LSTM &  0.89 $\pm$ 0.01  &  0.87 $\pm$ 0.01  &  0.83 $\pm$ 0.03  &  0.78 $\pm$ 0.03 & 0.53 / 0.59 & 0.266 / 0.009 \\
		\midrule
		CNN1D + LSTM &  0.80 $\pm$ 0.03  &  0.79 $\pm$ 0.02 &  0.69 $\pm$ 0.05  &  0.75 $\pm$ 0.07 & 0.33 / 0.4 & 0.009 / 0.003 \\
		CNN1D + BLSTM &  0.80 $\pm$ 0.01  &  0.78 $\pm$ 0.01 &  0.73 $\pm$ 0.03  &  0.71 $\pm$ 0.03 & 0.56 / 0.51  & 0.021 / 0.004 \\
		\midrule
        ResNet18 & 0.81 $\pm$ 0.02 & 0.78 $\pm$ 0.03 & 0.75 $\pm$ 0.04 & 0.75 $\pm$ 0.02 & 0.3 / 0.34 & 1.446 / 0.040   \\
		MobileNetV3 & 0.85 $\pm$ 0.01 & 0.83 $\pm$ 0.01 & 0.77 $\pm$ 0.04 & 0.76 $\pm$ 0.01 & 0.47 / 0.55 & 2.588 / 0.055   \\
		AlexNet & 0.83 $\pm$ 0.02 & 0.81 $\pm$ 0.02 & 0.70 $\pm$ 0.03 & 0.81 $\pm$ 0.02 & 0.56 / 0.64 & 0.079 / 0.015   \\
		DenseNet & 0.82 $\pm$ 0.01 & 0.79 $\pm$ 0.01 & 0.78 $\pm$ 0.04 & 0.73 $\pm$ 0.03 & 0.37 / 0.45 & 0.356 / 0.043   \\
        ChronoNet & 0.59 $\pm$ 0.03 & 0.57 $\pm$ 0.02 & 0.58 $\pm$ 0.18 & 0.56 $\pm$ 0.14 & 0.2 / 0.34 & 0.0136 / 0.0061   \\
		TDNN + LSTM & 0.80 $\pm$ 0.02 & 0.78 $\pm$ 0.02 & 0.72 $\pm$ 0.07 & 0.73 $\pm$ 0.05 & 0.39 / 0.47 & 3.49 / 0.24    \\
		\midrule
		Feature Transformer & 0.6 $\pm$ 0.13 & 0.6 $\pm$ 0.12 & 0.46 $\pm$ 0.24 & 0.72 $\pm$ 0.16 &  0.08 / 0.13 & 0.1231 /0.0074 \\
		Guided Feature Transformer  & 0.82 $\pm$ 0.01 & 0.81 $\pm$ 0.01 & 0.7 $\pm$ 0.03 & 0.8 $\pm$ 0.163 & 0.49 / 0.55 & 0.1646 / 0.0072 \\
		\bottomrule
	\end{tabular}
\end{table*}

In this section, we report comparative experimental result on 15 model architectures (Table \ref{Result}, Appendix Table \ref{unipolar}), 6 signal feature extractors (Table \ref{feature}), and 4 evaluation metrics (Table \ref{eval}). We first evaluate the performance and speed of 15 models (Table \ref{Result}) to find out the most favorable model for the real-time seizure detection task. We used the raw input feature as only the raw input showed the fastest processing speed on both CPU and GPU (Table \ref{feature}), among all the other signal feature extractors. As model evaluation only used raw signal, we explored other well-known signal feature extractors so as to achieve a better performance given the CNN2D+LSTM model. We also explored four different evaluation metrics previously used for EEG detection task, as Table \ref{Result} only reports EPOCH and MARGIN. We used CNN2D+LSTM among the 15 models for comparing signal feature extractors and evaluation metrics as this model provided accurate predictions as well as fast inference speed. We also provide the same set of experiments with ResNet-short+LSTM, which achieved the highest performance on raw EEG signals with acceptable real-time seizure detection inference speed (Table \ref{Result}), for cross-reference (Table \ref{windowsize_resnetlstm}, \ref{shift_resnetlstm}, \ref{eval_preproc_appendix_resnetlstm}).


\subsection{Real-Time Seizure Detection}
\label{sec:results-realtime}
We compare the performance of real-time binary seizure detectors on both bipolar (Table \ref{Result}) and unipolar lead (Appendix Table \ref{unipolar}) EEG signal datasets. We used a window size of 4 seconds and a shift length of 1 second after extensively searching for optimal window and shift size (Appendix Section \ref{sec:windowshift}).
ResNet-short+LSTM and ResNet-short + Dilation + LSTM both showed the best performance among various models (Table \ref{Result}). 
ResNet-short+LSTM performed better than the ResNet18 model as the LSTM module can capture the temporal dependency of the input signal.
However, the processing speed on CPU exceeds or similar to the shift length (1 second for the experiments in Table \ref{Result}) indicating that larger models are relatively slow. 

Compared to CNN1D+LSTM, CNN2D+LSTM showed better performance as it considers independently extracted channel-wise feature information.
ChronoNet \citep{roy2019chrononet} performed the worst probably because we used the raw signal and did not follow the feature extractor (LFCC) from the original paper for fair comparison. Also, the model was originally designed for detecting abnormal signals (seizure or abnormal noisy signals) which is rather easier task compared to seizure detection.
Feature Transformer also showed severely poor performance but gained significant performance increase after we applied the channel adjacency mask (\textit{i.e.} Guided Feature Transformer).
Thus, channel relation mask helps the model factor in similarity between channels on top of attention scaling, so that model can capture a seizure signal spreading out to the neighborhood leads.

Model performance on the Unipolar dataset showed similar trends as in the Bipolar dataset (Appendix Table \ref{unipolar}), although models suffered slight decrease in performance compared to the Bipolar dataset.
This might be generally because Bipolar montage is effective in localizing and lateralizing seizure \citep{beniczky2020electroencephalography}. 

CNN2D + LSTM model performed on par with ResNet18 based models, and was fast enough to meet the real-time constraint (\textit{i.e.} a model must process each window within shift length) (Table \ref{Result}). Taking into account the trade-offs between model size (Appendix Table \ref{modelsize}) and performance, we used CNN2D+LSTM model for further evaluation on bipolar dataset with various evaluation methods (Table \ref{eval}), window size (Appendix Table \ref{windowsize}), shift size (Appendix Table \ref{shift}), signal feature extractors (Appendix Table \ref{feature}), and seizure-type binary detection (Appendix Table \ref{seizuretype}). 

\begin{table*}[ht!]
	\footnotesize
	\centering
	\caption{Real-time seizure detection on bipolar TUH EEG dataset trained with each different signal process feature extractor on CNN2D + LSTM, averaged over $5$ runs. Please see Appendix section \ref{sec: speed} for CPU and GPU settings. Detailed results can be found in Appendix Table \ref{eval_preproc_appendix}.}
	\label{feature}
	\begin{tabular}{c|cccccc}
		\toprule
		Methods & AUROC & AUPRC & TPR & TNR & MARGIN(5sec)  &  CPU/GPU\\
		&&&&& Acc(Onset / Offset) & Speed (sec)\\
		\midrule
		\textbf{Raw} 
		& 0.89 $\pm$ 0.01
		& 0.88 $\pm$ 0.01 
		&  0.81 $\pm$ 0.03  
		&  0.83 $\pm$ 0.03
		& \textbf{0.56} / 0.51 
		& \textbf{0.079 / 0.004} \\
		SincNet 
		& 0.83 $\pm$ 0.01 
		& 0.81 $\pm$ 0.02 
		& 0.72 $\pm$ 0.03 
		& 0.78 $\pm$ 0.01 
		& 0.5 / 0.52 
		& 0.293 / 0.017 
		\\
		\textbf{STFT} 
		& \textbf{0.91 $\pm$ 0.01 }
		& \textbf{0.90 $\pm$ 0.01}
		& \textbf{0.85 $\pm$ 0.03 }
		& 0.82 $\pm$ 0.03 
		& 0.55 / \textbf{0.59} 
		& 0.538 / 0.167 
		\\
		
		\textbf{Frequency Bands} 
		& \textbf{0.92 $\pm$ 0.01}  
		& \textbf{0.91 $\pm$ 0.01} 
		& \textbf{0.85 $\pm$ 0.02} 
		& 0.83 $\pm$ 0.01 
		& \textbf{0.56} / 0.57 
		& 0.131 / 0.254 
		\\
		
		LFCC & 0.86 $\pm$ 0.03 & 0.84 $\pm$ 0.024 & 0.73 $\pm$ 0.07 & 0.84 $\pm$ 0.03 & 0.52 / 0.51 & \textbf{0.051} / 0.301 \\
		Downsampled Raw & 0.88 $\pm$ 0.02 & 0.87 $\pm$ 0.02 &  0.78 $\pm$ 0.04 &  0.83 $\pm$ 0.03 & 0.53 / 0.55 & 0.143 / \textbf{0.006} \\
		\bottomrule
	\end{tabular}
\end{table*}
\subsection{Seizure Detection with various signal processing methods}
We compared the performance and speed of traditionally and frequently used signal processing methods (Table \ref{feature}). Frequency Bands and STFT showed better performance compared to raw feature extraction, and SincNet \citep{ravanelli2018speaker} showed the worst performance among all feature extractors. The raw signal showed the fastest CPU/GPU processing speed with CNN2D+LSTM, and showed similar or better MARGIN and TAES performance compared to frequency bands and STFT with both CNN2D+LSTM and Resnet-short+LSTM models (Appendix Table \ref{eval_preproc_appendix}, \ref{eval_preproc_appendix_resnetlstm}). We used raw data for data preparation as this provides original data without any information loss, and maintains fast processing speed suitable to be used in conjunction with EEG test in clinical setting.

\begin{table*}[ht!]
	\footnotesize
	\centering
	\caption{Exploration on four evaluation methods on real-time seizure detection with CNN2D+LSTM, average over 5 runs. We selected TPR, TNR and FAs/24hours that maximizes TPR + TNR, and selected MARGIN and Onset Latency when TNR is above 0.95.}
	\label{eval}
	\begin{tabular}{c|c|ccccccc}
		\toprule
		& Metrics & AUROC & AUPRC & TPR & TNR & FAs / 24 hrs & Acc(Onset, Offset) & Time(Sec)\\
		\midrule
		\multirow{6}{*}{CNN2D+LSTM} 
		& OVLP & - & - & 0.75 & 0.82 & 47.06 & - & - \\
		& TAES & - & - & 0.33 & 1.0 & 1.03 & - & - \\
		& EPOCH & 0.89 & 0.88 & 0.81 & 0.83 & - & - & - \\
		& MARGIN(3sec)  & - & - & - & - & - & 0.41, 0.5 & - \\
		& MARGIN(5sec) & - & - & - & - & - & 0.56, 0.51 & - \\
		& Onset Latency & - & - & - & - & - & - & 10.55 \\
		\bottomrule
	\end{tabular}
\end{table*}

\subsection{Seizure Detection Evaluation Metrics}
We then evaluated the binary seizure detection with CNN2D+LSTM and Resnet-short+LSTM using various evaluation methods (Table \ref{eval}, \ref{eval_preproc_appendix}, \ref{eval_preproc_appendix_resnetlstm}). Sensitivity (TPR) measured with OVLP was the highest among all evaluation metric as it is more permissive in terms of sensitivity than other evaluation metrics \citep{ziyabari2017objective}.

We also present our novel seizure evaluation method MARGIN (seizure boundary detection) on onset/offset considering margin from the onset/offset time. Our example patient signal streak shows that MARGIN provides additional assessment on the relibility of a model whether the real-time detection output near the onset/offset of seizure is trustworthy or not (Figure \ref{fig_eval}). 

In this sample patient EEG trajectory (Figure \ref{fig_eval}) and from the result (Table \ref{eval}), OVLP showed high performance as it provides loose evaluation on the overlapping signal duration between ground truth and prediction. On the other hand, TAES provides stricter result as it considers overlapping time between label and hypothesis. Onset latency itself could not also be a favorable metric for real-time seizure event detection as we need to further consider the total length of seizure events (Appendix Table \ref{length}) which is is not possible to be done real-time.

EPOCH presents evaluation result on every windows, temporally detecting unreliable prediction of a model. We used EPOCH for other experiments as it has been used in previous literature and provides the strict, per epoch evaluation. EPOCH might be of concern if the dataset contains subjects presenting with long and short seizure events \citep{ziyabari2017objective}, as it weigh long seizure events heavily. We thus provide additional metric more useful for seizure onset detection (MARGIN). 
\section{Conclusion}
\label{sec:conclusion}
We provide a comprehensive analysis of the experimental settings on binary EEG seizure detection task for future benchmarking. While each existing models were successful in detecting seizures, the tasks cannot be compared against each other as the tasks were distinctly designed and tested. We set a standard real-time seizure detection task to fairly compare each models and experimental settings. We validate models on real-time binary seizure detection tasks with raw sampled signal under the constraint that each window signal is processed within shift length. This work could be extended to multi-class seizure classification tasks and localization of seizure onset with the help of attention modules. Our extensive research would help researchers initiate research in EEG based seizure detection tasks.

\section{Institutional Review Board (IRB)}
Our work does not require Institutional Review Board approval as we did not collect patient dataset for our research.

\acks{
 This work was supported by Institute of Information \& Communications Technology Planning \& Evaluation (IITP) grant (No.2019-0-00075, Artificial Intelligence Graduate School Program(KAIST)) and National Research Foundation of Korea (NRF) grant (NRF-2020H1D3A2A03100945), funded by the Korea government (MSIT).
}

\bibliography{reference}

\clearpage
\appendix
\section{Detailed information on Experimental Setting}

\subsection{Dataset}
\label{sec: app_dataset}
\begin{table*}[h!]
    \footnotesize
	\centering
	\caption{Construction of Temple University Hospital v 1.5.2 Dataset}
	\label{tuh}
	\begin{tabular}{c|cccc}
		\toprule
		Type & Patient No. & Seizure Events & Seizure Events (Train set) & Seizure Events (Development set) \\
		\midrule
		Focal Non-Specific & 150 &  1,836  & 1,536 & 300 \\
		Generalized Non-Specific  & 81 & 583 & 409 & 174 \\
		Simple Partial Seizure & 3 & 52 & 49 & 3 \\
		Complex Partial Seizure   & 41 & 367 & 283 & 84 \\
		Absence Seizure  & 12  & 99 & 50 & 49 \\
		Tonic Seizure  & 3  & 62 & 18 & 44 \\
		Tonic Clonic Seizure  & 14 & 48 & 30 & 18 \\
		\bottomrule
	\end{tabular}
\end{table*}

We slightly modified the given TUH V1.5.2 data in order to detect each seizure types with ample number of signals. We merged the original Tonic Seizure (TNSZ) and Tonic-Clonic Seizure (TCSZ) in TUH-EEG Corpus according to $2017$ ILAE guideline \citep{scheffer2017ilae}. Also, we removed Myoclonic Seizure (MYSZ) seizure type data since there are only two patients, and transferred one SPSZ patient and one TNSZ patient data to test data from original TUH train data in order to allow all train, validation, and test set to have equal number of patient data for each type of seizure. 

Normal signals in TUH dataset includes when a patient presents with Eye Movement (EYEM), Chewing (CHEW), Shivering (SHIV), muscle artifact (MUSC), Electrode Pop (ELPP), and Electrostatic Artifact (ELST). Among $7,034$ of seizure signals, $583$ in Focal Non-Specific seizure (FNSZ) class, $1,836$ signal streaks in Generalized Seizure (GNSZ),  $52$ in Simple Partial Seizure (SPSZ), $367$ in Complex Partial Seizure (CPSZ), $99$ in Absence Seizure (ABSZ), $62$ in Tonic Seizure (TNSZ), and $48$ in Tonic Clonic Seizure (TCSZ) (Table \ref{tuh}). 

Based on the original label from the original EEG, we re-defined the label for each input window. If a signal window contains ictal period of more than the length of window shift length, we set the signal window is \textit{ictal}, and \textit{nonictal}, otherwise.

\begin{figure}[h!]
	\centering
	\includegraphics[width=\columnwidth]{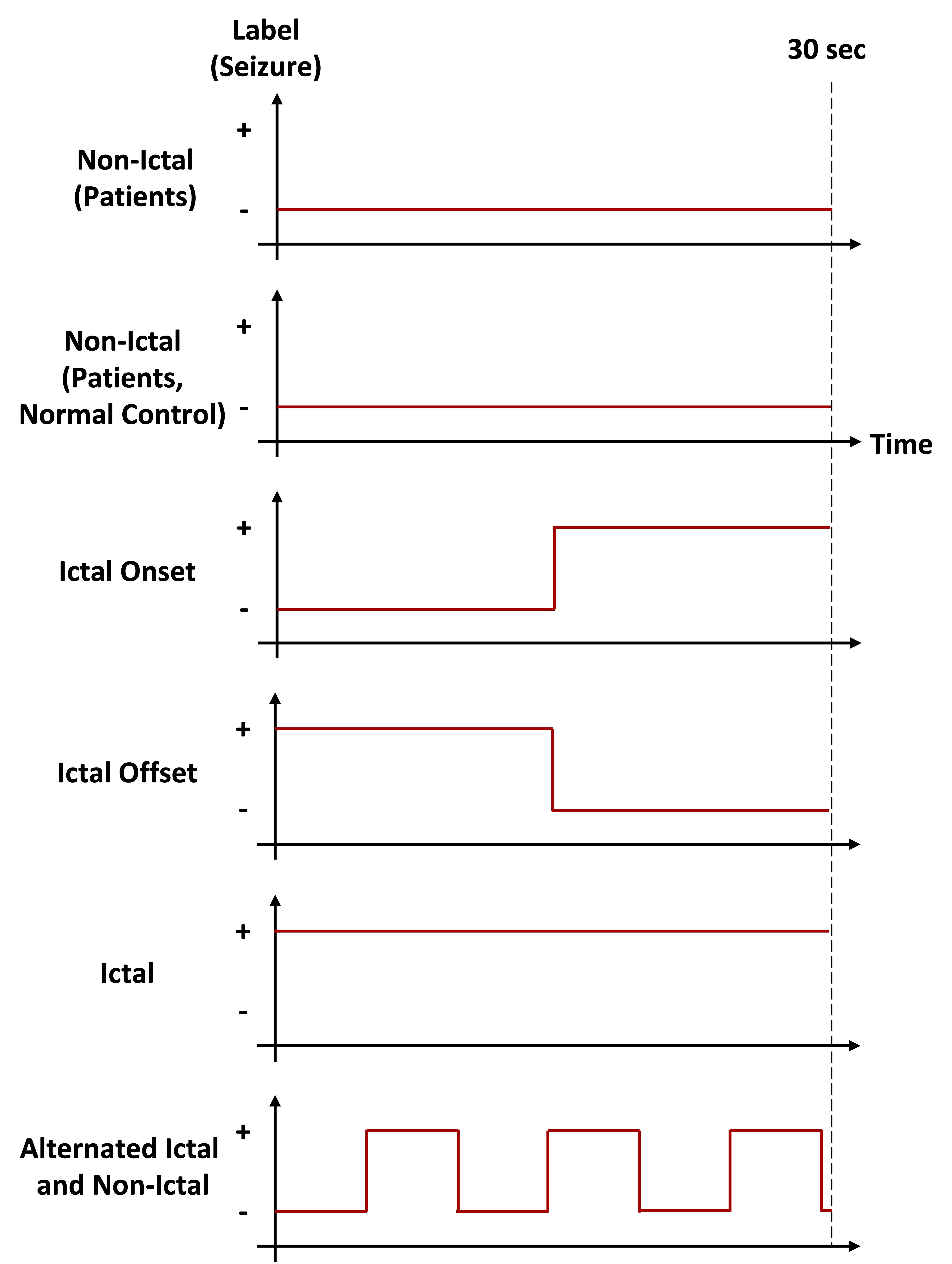}
    \caption{\small \textbf{Data Sampler Inputs} We equally distribute Non-Ictal, Ictal Onset/Offset, Ictal, and Alternated Ictal/Non-Ictal signal segments within each batch.}
	\label{sampling}
\end{figure}

\begin{figure*}[h!]
	\centering
	\includegraphics[width=\textwidth]{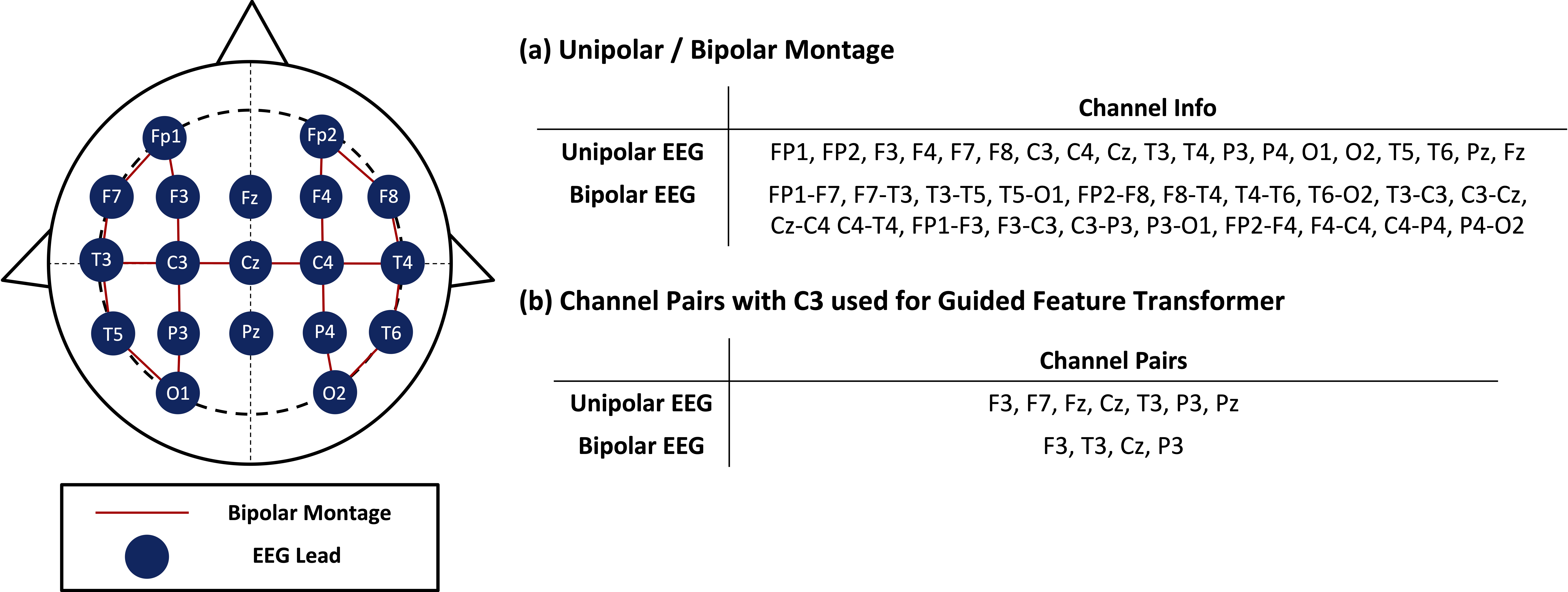}
    \caption{\small \textbf{Lead Information for TUH Dataset} We summarize the channels in unipolar and bipolar montage (left figure and Table (a) Unipolar / Bipolar Montage). We used the neighborhood leads for Unipolar Montage and interconnected Bipolar channels to generate adjacency matrix in Guided Feature Transformer (Figure \ref{featatt}). Table (b) examplifies the channel pairs with C3 for Unipolar and Bipolar Montage, respectively.}
	\label{leadinfo}
\end{figure*}
\begin{table}[h!]
    \footnotesize
	\centering
	\caption{Average length of seizure events within EEG signals of each seizure types.}
	\label{length}
	\begin{tabular}{c|c}
		\toprule
		Type & Seizure Event (sec) \\
		\midrule
		Background & 357.36 $\pm$ 384.29\\
		Focal Non-Specific & 65.98 $\pm$ 143.55\\
		Generalized Non-Specific & 102.43 $\pm$ 265.18\\
		Simple Partial Seizure & 41.27 $\pm$ 19.81\\
		Complex Partial Seizure & 98.97 $\pm$ 112.39\\
		Absence Seizure & 8.61 $\pm$ 5.34\\
		Tonic Seizure & 19.41 $\pm$ 9.30\\
		Tonic Clonic Seizure & 115.59 $\pm$ 327.95\\
		\bottomrule
	\end{tabular}
\end{table}
We also summarize the length of the seizure events for each seizure types in Table \ref{length}. Background noise (normal signal types) showed the longest event duration, and abscence seizure showed the shortest event length. Tonic Clonic Seizure showed the longest event duration among all seizure types.

\subsection{Signal Feature Extractor}
\label{sec: featextractor}
\textbf{1) Raw}
$64 \times 1 \times 51$ CNN2D filter is convolved to the raw input signal with 20 channels and 800 data points across the time dimension for channel-wise feature extraction.

\noindent\textbf{2) SincNet}
We used 7 distinct SincNet filters of size $1 \times 80$ to extract information from each signal frequency. Once the SincNet filters are convoluted with input EEG signal, the output size becomes $7 \times 20 \times 400$, 7 different frequency information of length 400 for all 20 channels. It is then flattened to $1 \times 140 \times 400$ so that every 7 row corresponds to 7 different frequencies per each channel. 
We designed the first CNN layer for SincNet to have the filter size of $64 \times 7 \times 21$ and stride size $7 \times 2$ for channel-wise feature extraction.

\noindent\textbf{3) Frequency bands}
The output of frequency bands is $20 \times 7 \times 100$, 7 averaged frequency bands of length 100 for each channel. This output is flattened to $1 \times 140 \times 100$ so that every 7 row corresponds to frequency band information for each channel. We used the first CNN layer with filter size $64 \times 7 \times 21$ and stride size $7 \times 2$ for channel-wise convolution.

\noindent\textbf{4) STFT}
Output of STFT ($20 \times 100 \times 100$) includes information of 100 frequencies of length 100 for all 20 channels. The output is flattened to size $1 \times 2,000 \times 100$ so that every row holds information about 100 frequencies per each channel. We also used CNN layer with filter size $64 \times 100 \times 3$ and stride $100 \times 1$ for channel-wise convolution.

\noindent\textbf{5) LFCC}
Output of LFCC ($20 \times 8 \times 27$) includes information of 8 size frequency domain for all 20 channels. The output is flattened to size $1 \times 160 \times 27$ so that every row holds information about 8 frequencies per each channel. We also used CNN layer with filter size $64 \times 8 \times 21$ and stride $8 \times 2$ for channel-wise convolution.

\noindent\textbf{6) Down-sampled}
The 800 data points of raw data is re-sampled in 200, 100, 50Hz then converted to size of 800, 400, and 200 data points samples. Each sample goes to different stride size CNN2D modules (stride size of 1x4, 1x2, 1x1) and outputs of the three CNN2D modules are concatenated to size $64 \times 20 \times 200$. We additionally implement and compare experiments of Down-sampled feature extraction with and without anti-aliasing filter (bandpass filter) (Table \ref{antialiasing}). The anti-aliasing filter does not improve the performance of our model.

\subsection{Model Architectures}
\label{sec: app_arch}
\begin{figure}[h!]
	\centering
	\includegraphics[width=\columnwidth]{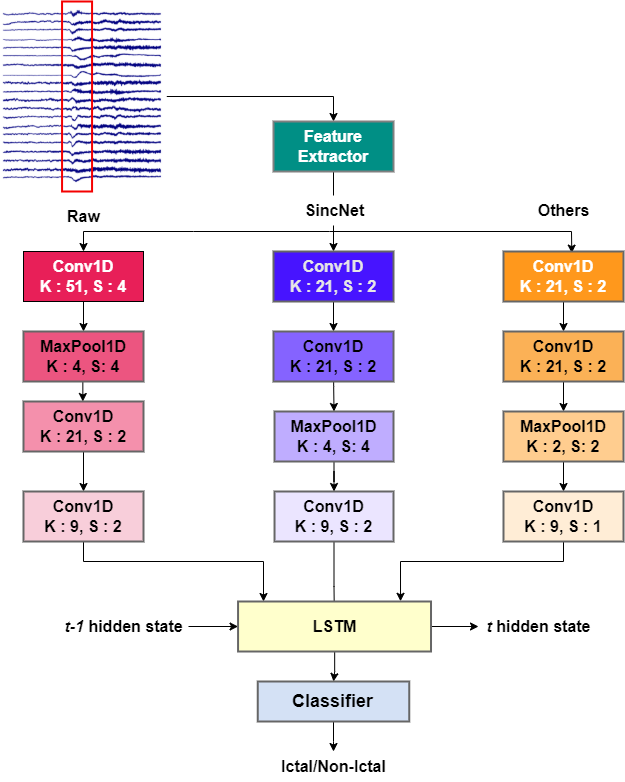}
    \caption{\small \textbf{Architecture of CNN1D + LSTM seizure detector} according to each feature extractors. K: Kernel size, S: Stride}
	\label{1D_architecture}
\end{figure}
\begin{figure*}[h!]
	\centering
	\includegraphics[width=0.9\textwidth]{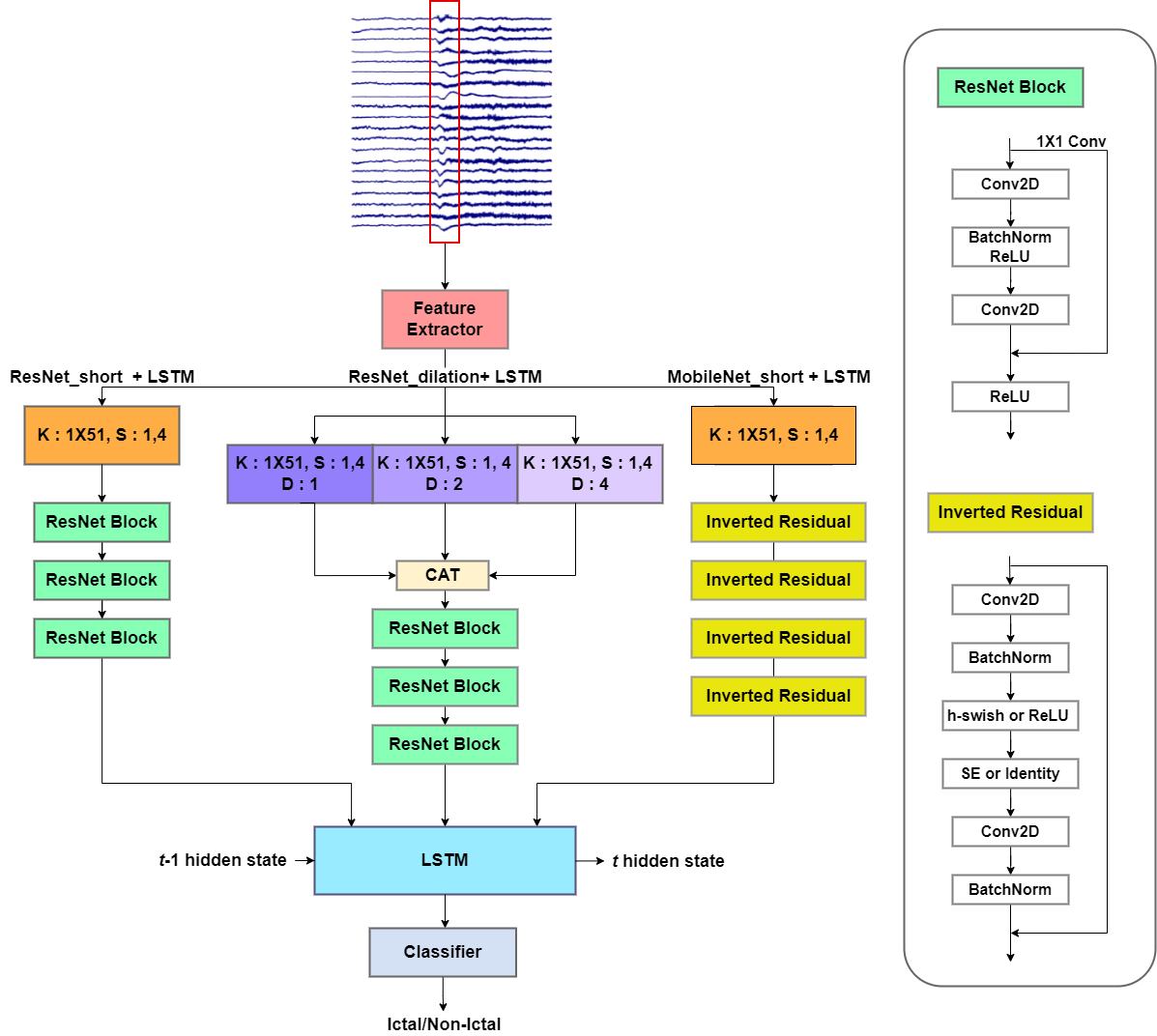}
    \caption{\small \textbf{Architecture of ResNet-short+LSTM, ResNet-short+Dilation+LSTM, and MobileNetV3-short+LSTM Models} K: Kernel size, S: Stride}
	\label{resnetmobile}
\end{figure*}
\begin{figure*}[h!]
	\centering
	\includegraphics[width=\textwidth]{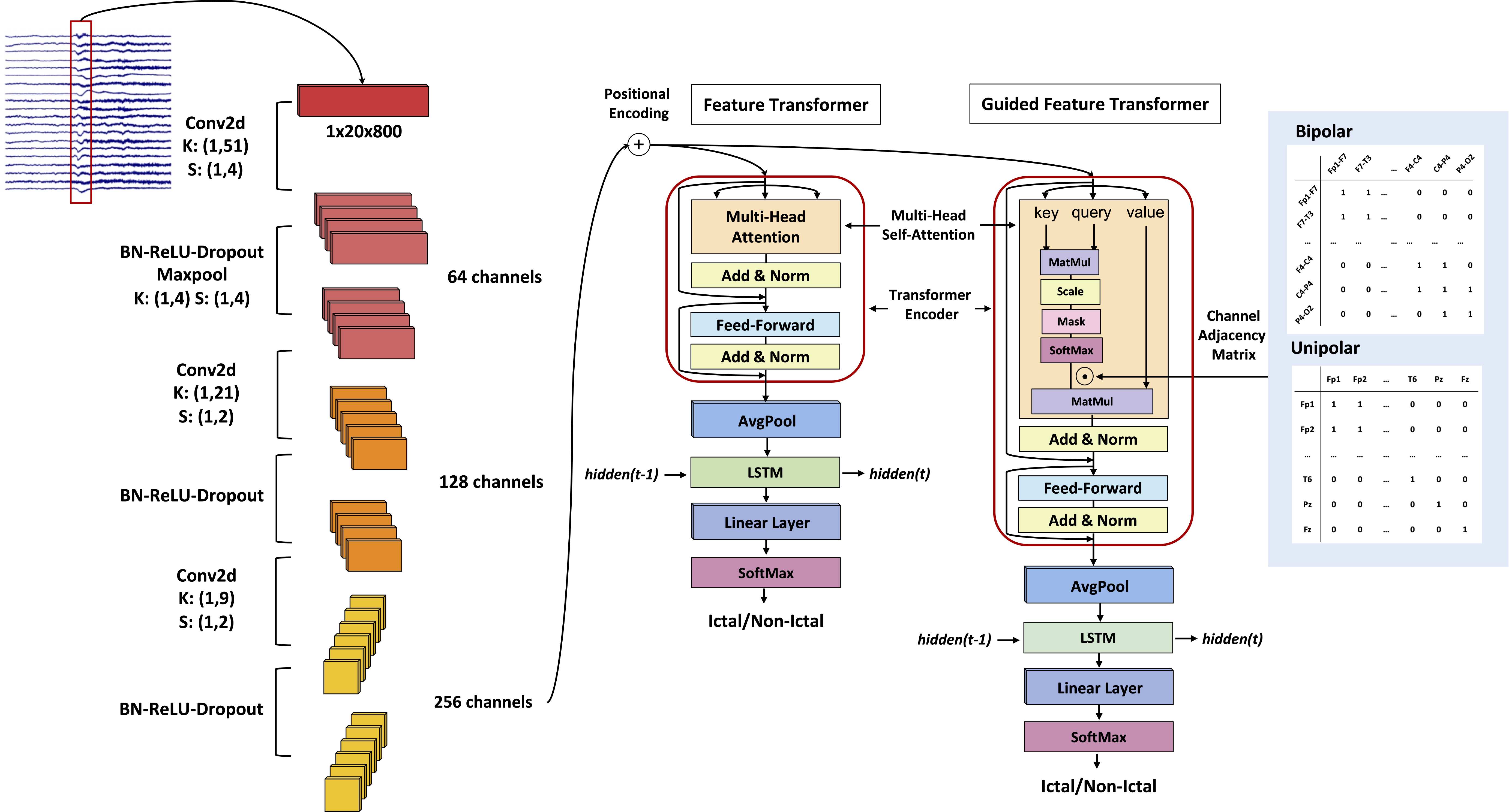}
    \caption{\small \textbf{Architecture of Feature Transformer and Guided Transformer}. Feature Transformer and Guided Transformer shares same architecture, but the channel adjacency matrix is multiplied to the key and query output of multi-head self attention module in Guided Transformer transformer encoder.}
	\label{featatt}
\end{figure*}
We summarized the CNN1D architecture (Figure \ref{1D_architecture}), ResNet and MobileNetV3 based CNN2D+LSTM models (ResNet-short+LSTM, ResNet-short+Dilation+LSTM, and MobileNetV3-short+LSTM model, Figure \ref{resnetmobile}), and feature transformers (Figure \ref{featatt}). 

\noindent\textbf{CNN1D+LSTM} (Figure \ref{1D_architecture}): Upon feature extraction, the feature passes CNN1D layers and MaxPool layer, LSTM layer and then linear layer for binary classification. 

\noindent\textbf{ResNet/MobileNetV3-based CNN2D+LSTM} (Figure \ref{resnetmobile}): ResNet\_short+LSTM and ResNet\_short+ Dilation+LSTM models both have ResNet blocks from the original ResNet18 encoder which feed output to LSTM and linear classifier. ResNet\_short+Dilation+LSTM model additionally have dilation and concatenation before ResNet block. MobileNet\_short+LSTM layer have Inverted Residual blocks from original MobileNetV3 architecture, which also feeds the output into LSTM and linear classifier.

\noindent\textbf{Feature Transformer, Guided Feature Transformer} (Figure \ref{featatt}): The EEG input signal first passes the CNN filter and then positional encoding was to time dimension and then passed to transformer encoder \citep{vaswani2017attention}. In Guided Feature Transformer, the channel adjacency matrix is multiplied element-wise to the sofmax output of key query of multi-head self attention layer to calculate the similarity between connected channels. Transformer encoder output then passes through average pooling, LSTM, and linear classifier layer for seizure binary detection.

\subsection{Model Speed}
\label{sec: speed}
GPU and CPU speed were measured with Titan Xp and Intel(R) Xeon(R) CPU E5-2640 v4 @ 2.40GHz. We measured both inference and learning time of each signal feature extraction methods with CNN2D+LSTM model on raw signal. The inference time was calculated per each sliding window on one CPU, where we measured the total time it took to infer a $30$ second input signal segment. We report the total time divided by the number of windows ($4$ second window, $26$ windows) within the whole $30$ second signal. 

\noindent\textbf{1) CPU Processing Speed}: CPU speed is the time it takes for a model to process one sliding window using only one CPU with batch size 1. This is the realistic model evaluation in a hospital where the EEG data from one patient is processed by one microprocessor.

\noindent\textbf{2) GPU Processing Speed}: GPU speed is the time it takes for a model to process one sliding window using one GPU with the support of 5 CPUs.

\section{Supplementary Results}
\subsection{Model Size}
\label{sec: size}
We summarize the model size according to each model (Table \ref{modelsize}). Chrononet and TDNN + LSTM are the two smallest models; AlexNet is the biggest model, followed by the ResNet based model. The sizes of the models were all measured using torchinfo API.
\begin{table}[ht!]
	\footnotesize
	\centering
	\caption{\textbf{Model Configurations. } Model parameter size measured with torchinfo API.}
	\label{modelsize}
	\begin{tabular}{c|c}
		\toprule
		Methods  & Parameter Size (MB) \\
		\midrule
		CNN2D + LSTM &	6.16   \\
		CNN2D + BLSTM & 6.16 \\
		ResNet-short + LSTM & 15.75  \\
		ResNet-short + Dilation + LSTM & 15.75  \\
		MobileNetV3-short + LSTM & 5.44   \\
		\midrule
		CNN1D + LSTM & 6.41   \\
		CNN1D + BLSTM & 7.99  \\
		\midrule
		ResNet18  	&20.82 \\
		MobileNetV3 & 	16.54   \\
		AlexNet	& 127.86  \\
		DenseNet &	4.84   \\
		ChronoNet & 0.5 \\
		TDNN + LSTM & 2.67 \\
		\midrule
		Feature Transformer + LSTM & 5.6\\
		Guided Transformer + LSTM &  5.6\\
		\bottomrule
	\end{tabular}
\end{table}

\subsection{Seizure Detection with Multiple Window Size and Shift Length}
\label{sec:windowshift}
We evaluate the performance of CNN2D+LSTM and ResNet-short+LSTM using various input window size (1, 2, 4, 6, 7, 8, 10, 12 sec) (Table \ref{windowsize}, \ref{windowsize_resnetlstm}) and window shift length (1, 2, 3, 4, 5 sec) (Table \ref{shift}, \ref{shift_resnetlstm}). The model performs similar regardless of the window size, and we got the best performance with the largest window size ($12$ second). With increased window size, we can utilize more information within the input signal but at the same time, the processing speed decreases gradually (increasing processing time in Table \ref{windowsize}) and also might be inconvenient to users as they have to wait for the device to gather information for the time equivalent to the window size. Also, wider window size decreases the temporal resolution of the seizure detection alarm raised by the real-time seizure detector as the model detects the seizure only after it acquires the signal within the full window. We used the $4$ second window setting for other experiments as this setting shows the second-best performance and still includes some input signal information than shorter window settings.
\begin{table}[h!]
    \small
	\centering
	\caption{Real-time seizure detection with different \textbf{window size} on  bipolar TUH EEG dataset trained with CNN2D+LSTM, averaged over $5$ runs.The CPU speed setting and meaning are explained in \ref{sec: speed}}
	\label{windowsize}
	\begin{tabular}{c|ccc}
		\toprule
		Window & AUROC & AUPRC & Process\\
		(sec) &&& (sec)\\
		\midrule
		1 & 0.87 $\pm$ 0.02 & 0.86 $\pm$ 0.02 & \textbf{0.0364} \\
		2  & 0.88 $\pm$ 0.02 & 0.87 $\pm$ 0.02  & \textbf{0.0665} \\
		\textbf{4} & 0.89 $\pm$ 0.01 & 0.88 $\pm$ 0.01 & \textbf{0.079}\\
		6  & 0.89 $\pm$ 0.02 & 0.88 $\pm$ 0.02 & 0.1243 \\
		7  & 0.88 $\pm$ 0.02 & 0.87 $\pm$ 0.02 & 0.1305 \\
		8  & 0.88 $\pm$ 0.02 & 0.87 $\pm$ 0.02 & 0.1222 \\
		10  & 0.89 $\pm$ 0.01 & 0.87 $\pm$ 0.01 & 0.2441 \\
		12  & 0.90 $\pm$ 0.01 & 0.88$\pm$ 0.01  & 0.2616 \\
		\bottomrule
	\end{tabular}
\end{table}

\begin{table}[h!]
	\small
	\centering
	\caption{Real-time seizure detection with different \textbf{window shift length} on bipolar TUH EEG dataset trained with CNN2D+LSTM with fixed 4 second sliding window, averaged over $5$ runs.}
	\label{shift}
	\begin{tabular}{c|ccc}
		\toprule
		Shift Length & AUROC & AUPRC \\
		(sec)\\
		\midrule
		1 & \textbf{0.89 $\pm$ 0.01} & \textbf{0.88 $\pm$ 0.01}\\
		2 & 0.87 $\pm$ 0.05 & 0.86 $\pm$ 0.04\\ 
		3 & 0.86 $\pm$ 0.04 & 0.86 $\pm$ 0.04\\
		4 & 0.85 $\pm$ 0.04 & 0.84 $\pm$ 0.04\\
		5 & 0.84 $\pm$ 0.05 & 0.84 $\pm$ 0.05\\
		\bottomrule
	\end{tabular}
\end{table}

The model showed worse performance as the shift size increases. As shift length increases model process speed increases, but a model can provide more inference results by shifting the window every second. We used the best performing shift length (1 second) for our final setting as it can utilize and learn more windows than any other setting, closer to a real-world setting, and provides the highest performance.

\subsection{Model Speed}
\begin{table*}[h!]
    \footnotesize
	\centering
	\caption{Computation speed (sec) per one 4 seconds size window of CNN2D + LSTM (RAW) as base model, Resnet-short + LSTM (RAW) as deeper nerual network model and CNN2D + LSTM (Frequency Based) as more complex signal feature extraction method model according to input batch size.}
	\label{speed_batch}
	\begin{tabular}{c|c|cccccc}
		\toprule
		Model (Feature) & GPU/CPU & & & Batch Size &  & & \\
		&& 1 & 4 & 8 & 16 & 32 & 64 \\
		\midrule
		CNN2D + LSTM & 1 CPU & 0.0792 &	0.3031 & 0.571 &	1.1527 & 2.27 &	4.4862  \\
		(Raw) & 1 GPU + 5 CPUs & 0.0036 & 0.0028 & 0.0026 & 0.0034 & 0.0043 & 0.0069\\
		\midrule
		ResNet-short + LSTM & 1 CPU & 0.941 & 4.0771 & 7.741 & 16.4739 & 33.2041 & 67.0107 \\
		(Raw) & 1 GPU + 5 CPUs & 0.013 & 0.0049 & 0.0056 & 0.0077 & 0.013 & 0.0213\\
		\midrule
		CNN2D + LSTM & 1 CPU & 0.131 & 0.3403 & 0.6433 & 1.2804 & 2.4279 & 4.9421 \\
        (Frequency Bands) & 1 GPU + 5 CPUs & 0.0106 & 0.0344 & 0.0687 & 0.1354 & 0.254 & 0.6481\\
		\bottomrule
	\end{tabular}
\end{table*}
The batch size of EEG training input might affect the processing speed of each model. As batch size increases, the processing time of each STFT based signal feature extraction increases linearly to the number of EEG channels as it cannot be done in parallel within GPU processors. Thus, although CPU based on-device (1 batch size) speed doesn't have huge difference between raw and STFT feature based models, the STFT feature based model's training speed is much lower than the raw feature based neural network. We compared the processing speed of CNN2D + LSTM (Raw) as base model with Resnet-short + LSTM (RAW) as deeeper neural network model and CNN2D + LSTM (Frequency band) as complex signal feature extraction model according to batch size (Table \ref{speed_batch}).

\subsection{Real-time Seizure Detection on Unipolar EEG lead}
We trained 15 models on raw Unipolar EEG dataset and report the test result in Table \ref{unipolar}. Training on Unipolar dataset showed similar trends as Bipolar dataset, where ResNet-short variant of CNN2D+LSTM model showed the best performance and CNN2D+LSTM showed performance on par with ResNet based CNN2D+LSTM models.

\begin{table*}[ht!]
	\footnotesize
	\centering
	\caption{Result of real-time seizure detection on \textbf{raw Unipolar} TUH EEG dataset trained with each architecture, averaged over $5$ runs.}
	\label{unipolar}
	\begin{tabular}{c|cccc}
		\toprule
		Methods & AUROC & AUPRC & TPR & TNR\\
		\midrule
		CNN2D + LSTM & 0.84 $\pm$ 0.01 
		& 0.83 $\pm$ 0.01 
		& 0.71 $\pm$ 0.05 
		& \textbf{0.82 $\pm$ 0.05} 
		\\
		CNN2D + BLSTM 
		& 0.86 $\pm$ 0.03 & 0.85 $\pm$ 0.03 & 0.75 $\pm$ 0.03 
		& \textbf{0.81 $\pm$ 0.03 }
		\\
		ResNet-short + LSTM 
		& 0.86 $\pm$ 0.01 
		& 0.85 $\pm$ 0.02 
		& 0.77 $\pm$ 0.02 
		& 0.8 $\pm$ 0.03 
		\\
		
		ResNet-short +Dilation + LSTM 
		& \textbf{0.88 $\pm$ 0.01}
		& \textbf{0.87 $\pm$ 0.01}
		& 0.79 $\pm$ 0.01 
		& 0.8 $\pm$ 0.03
		\\
		
		MobileNetV3 + LSTM 
		& \textbf{0.87 $\pm$ 0.01 }
		& 0.85 $\pm$ 0.01 
		& \textbf{0.82 $\pm$ 0.01}
		& 0.75 $\pm$ 0.01
		\\
		
		\midrule
		CNN1D + LSTM & 0.8 $\pm$ 0.02  & 0.79 $\pm$ 0.03  & 0.66 $\pm$ 0.04 & 0.78 $\pm$ 0.04\\
		CNN1D + BLSTM & 0.78 $\pm$ 0.03 & 0.76 $\pm$ 0.03 & 0.71 $\pm$ 0.03  & 0.71 $\pm$ 0.03\\
		\midrule
		ResNet18 & 0.83 $\pm$ 0.01 & 0.80 $\pm$ 0.02 & 0.73 $\pm$ 0.03 & 0.77 $\pm$ 0.04\\
		
		MobileNetV3 
		& 0.85 $\pm$ 0.01 
		& 0.84 $\pm$ 0.01 
		& 0.78 $\pm$ 0.02 
		& 0.77 $\pm$ 0.05\\
		
		AlexNet 
		& 0.82 $\pm$ 0.02 
		& 0.81 $\pm$ 0.02 
		& 0.69 $\pm$ 0.05 
		& \textbf{0.83 $\pm$ 0.02}
		\\
		DenseNet 
		& 0.82 $\pm$ 0.02 
		& 0.79 $\pm$ 0.03 
		& 0.72 $\pm$ 0.04 
		& 0.77 $\pm$ 0.04
		\\
		
		ChronoNet & 0.63 $\pm$ 0.02 & 0.61 $\pm$ 0.02 & 0.63 $\pm$ 0.08 & 0.56 $\pm$ 0.08\\
		TDNN + LSTM & 0.70 $\pm$ 0.02 & 0.69 $\pm$ 0.02 & 0.56 $\pm$ 0.07 & 0.74 $\pm$ 0.04\\
		\midrule
		Feature Transformer & 0.51 $\pm$  0.02 & 0.51 $\pm$ 0.01  & 0.28 $\pm$ 0.24 &  0.75 $\pm$ 0.21\\
		Guided Feature Transformer  & 0.81 $\pm$ 0.01  & 0.8 $\pm$ 0.01  & 0.69 $\pm$ 0.03 
		& \textbf{0.81 $\pm$ 0.02}\\
		\bottomrule
	\end{tabular}
\end{table*}

\subsection{Seizure Type Binary Detection}
\begin{table*}[ht!]
	\small
	\centering
	\caption{\textbf{Seizure type-wise performance} of our real-time seizure detectors on \textit{raw} bipolar TUH EEG V1.5.2 dataset. The results were averaged over $5$ runs.}
	\label{seizuretype}
	\begin{tabular}{c|c|ccccc}
		\toprule
		Model (Feature Extractor) &Seizure Type & AUROC & AUPRC & TPR & TNR\\
		\midrule
		\multirow{7}{*}{CNN2D + LSTM}
		&Generalize Seizure 
		& 0.85 $\pm$ 0.01 
		& 0.82 $\pm$ 0.01 
		& 0.78 $\pm$ 0.06 
		& 0.77 $\pm$ 0.07 
		\\
		&Focal Non-Specific Seizure 
		& 0.92 $\pm$ 0.00 
		& \textbf{0.92 $\pm$ 0.01 }
		& 0.83 $\pm$ 0.04 
		& 0.85 $\pm$ 0.03 \\
		&Simple Partial Seizure 
		& 0.93 $\pm$ 0.03 
		& \textbf{0.92 $\pm$ 0.03 }
		& 0.83 $\pm$ 0.03 
		& \textbf{0.90 $\pm$ 0.01 }
		\\
		&Complex Partial Seizure 
		& 0.86 $\pm$ 0.03 
		& 0.81 $\pm$ 0.02 
		& 0.71 $\pm$ 0.02 
		& \textbf{0.90 $\pm$ 0.02 }
		\\
		(Raw) &Absence Seizure 
		& \textbf{0.97 $\pm$ 0.02 }
		& \textbf{0.90 $\pm$ 0.06 }
		& \textbf{0.96 $\pm$ 0.04 }
		& \textbf{0.92 $\pm$ 0.03 }
		\\
		&Tonic Seizure 
		& 0.85 $\pm$ 0.04 
		& 0.79 $\pm$ 0.04
		& 0.89 $\pm$ 0.02 
		& 0.76 $\pm$ 0.04 
		\\
		&Tonic-Clonic Seizure 
		& 0.90 $\pm$ 0.06 
		& 0.81 $\pm$ 0.13 
		& 0.86 $\pm$ 0.07 
		& 0.83 $\pm$ 0.08 
		\\
		\midrule
		\multirow{7}{*}{Resnet-short + LSTM }
		&Generalize Seizure 
		& \textbf{0.94 $\pm$ 0.00} 
		& \textbf{0.94 $\pm$ 0.01} 
		& 0.83 $\pm$ 0.03 
		& \textbf{0.90 $\pm$ 0.03 }
		\\
		&Focal Non-Specific Seizure 
		& 0.90 $\pm$ 0.00 
		& 0.88 $\pm$ 0.01 
		& 0.84 $\pm$ 0.04 
		& 0.81 $\pm$ 0.03 
		\\
		&Simple Partial Seizure 
		&\textbf{0.94 $\pm$ 0.04 }
		& 0.89 $\pm$ 0.08 
		& \textbf{0.91 $\pm$ 0.03 }
		& 0.83 $\pm$ 0.07 
		\\
		&Complex Partial Seizure 
		& \textbf{0.95 $\pm$ 0.02 }
		& \textbf{0.95 $\pm$ 0.02 }
		& 0.83 $\pm$ 0.04 
		& \textbf{0.94 $\pm$ 0.02 }
		\\
		(Raw)
		&Absence Seizure 
		& 0.92 $\pm$ 0.02 
		& 0.90 $\pm$ 0.01 
		& 0.80 $\pm$ 0.04 
		& \textbf{0.95 $\pm$ 0.03 }
		\\
		&Tonic Seizure 
		& 0.92 $\pm$ 0.04 
		& 0.79 $\pm$ 0.09
		& \textbf{0.92 $\pm$ 0.06 }
		& 0.84 $\pm$ 0.07 
		\\
		&Tonic-Clonic Seizure 
		& 0.84 $\pm$ 0.06 
		& 0.78 $\pm$ 0.07 
		& 0.85 $\pm$ 0.05 
		& 0.74 $\pm$ 0.11 
		\\
		\midrule
		\multirow{7}{*}{CNN2D + LSTM}
		& Generalize Seizure 
		& 0.95 $\pm$ 0.01
		& \textbf{0.94 $\pm$ 0.00}
		& 0.87 $\pm$ 0.03
		& 0.90 $\pm$ 0.02 
		\\
		&Focal Non-Specific Seizure 
		& 0.86 $\pm$ 0.01
		& 0.83 $\pm$ 0.02 
		& 0.85 $\pm$ 0.05 
		& 0.72 $\pm$ 0.05 
		\\
		&Simple Partial Seizure 
		& 0.80 $\pm$ 0.17 
		& 0.78 $\pm$ 0.16
		& 0.81 $\pm$ 0.19
		& 0.79 $\pm$ 0.31
		\\
		&Complex Partial Seizure 
		& 0.94 $\pm$ 0.01
		& 0.92 $\pm$ 0.03
		& 0.90 $\pm$ 0.03
		& 0.86 $\pm$ 0.02
		\\
		(Frequency Bands)
		&Absence Seizure 
		& 0.92 $\pm$ 0.02
		& 0.90 $\pm$ 0.02
		& 0.78 $\pm$ 0.02
		& \textbf{0.97 $\pm$ 0.01}
		\\
		&Tonic Seizure 
		& \textbf{0.98 $\pm$ 0.01} 
		& 0.92 $\pm$ 0.03
		& \textbf{0.99 $\pm$ 0.01}
		& 0.91 $\pm$ 0.02
		\\
		&Tonic-Clonic Seizure 
		& 0.87 $\pm$ 0.08
		& 0.83 $\pm$ 0.07
		& 0.95 $\pm$ 0.02
		& 0.77 $\pm$ 0.12
		\\
		\bottomrule
	\end{tabular}
\end{table*}
We report the binary detector model inference result with each seizure type binary detection task, on two different signal feature extractor settings (Table \ref{seizuretype}). The binary seizure type detection with CNN2D+LSTM model showed that the model performed well with AUC over 0.9 on detecting most seizure types (Table \ref{seizuretype}) except Tonic-Clonic Seizure task. Tonic-Clonic Seizure (TCSZ) detection task showed the AUROC of 0.84 (Table \ref{seizuretype}) which is lower than binary seizure detection performance of CNN2D+LSTM (Table \ref{Result}). TCSZ detection task showed better performance with other frequency bands (Table \ref{seizuretype}, Frequecy Bands row), as extracted signal features might capture the slowly evolving TCSZ better than the raw input. 

\subsection{Seizure Evaluation Metric}
\begin{table*}[ht!]
	\tiny
	\vspace{-25pt}
	\centering
	\caption{Exploration on four evaluation methods on real-time seizure detection with various models. We evaluate TPR, TNR and FAs/24hours that maximizes TPR + TNR, and measured MARGIN and Latency when TNR is above 0.95.}
	\label{eval_models}
	\begin{tabular}{c|c|cccccccc}
		\toprule
		& Metrics & AUROC & AUPRC & TPR & TNR & FAs / 24 hours & Acc(Onset / Offset) & Time(Sec) \\
		\midrule
		\multirow{6}{*}{CNN2D+BLSTM} 
		& OVLP & - & - & 0.7 & 0.86 & 47.18 & - & - \\
		& TAES & - & - & 0.3 & 1.0 & 2.06 & - & - \\
		& EPOCH & 0.87 & 0.86 & 0.76 & 0.85 & - & - & -  \\
		& MARGIN(3sec)  & - & - & - & - & - & 0.44 / 0.58 &   \\
		& MARGIN(5sec) & - & - & - & - & - & 0.57 / 0.64 & \\
		& Onset Latency & - & - & - & - & - & - & 13.69 \\

		\midrule
		\multirow{6}{*}{ResNet-short + LSTM} 
		& OVLP & - & - & 0.75 & 0.84 & 64.9 & - & - \\
		& TAES & - & - & 0.4 & 1.0 & 1.68 & - & - \\
		& EPOCH & 0.92 & 0.91 & 0.83 & 0.85 & - & - & -  \\
		& MARGIN(3sec)  & - & - & - & - & - & 0.49 / 0.58 & -  \\
		& MARGIN(5sec) & - & - & - & - & - & 0.62 / 0.65 & -  \\
		& Onset Latency & - & - & - & - & - & - & 15.23 \\

		\midrule
		\multirow{6}{*}{ResNet-short + Dilation} 
		& OVLP & - & - & 0.62 & 0.89 & 35.7 & - & - \\
		& TAES & - & - & 0.37 & 1.0 & 1.67 & - & - \\
		& EPOCH & 0.91 & 0.9 & 0.84 & 0.83 & - & - & -  \\
		& MARGIN(3sec)  & - & - & - & - & - & 0.43 / 0.5 &  - \\
		+ LSTM& MARGIN(5sec) & - & - & - & - & - & 0.58 / 0.61 & -  \\
		& Onset Latency & - & - & - & - & - & - & 7.76 \\

		\midrule
		\multirow{6}{*}{MobileNetV3 + LSTM} 
		& OVLP & - & - & 0.55 & 0.94 & 15.73 & - & - \\
		& TAES & - & - & 0.34 & 0.99 & 2.61 & - & - \\
		& EPOCH & 0.89 & 0.87 & 0.83 & 0.78 & - & - & -  \\
		& MARGIN(3sec)  & - & - & - & - & - & 0.44 / 0.51 & -  \\
		& MARGIN(5sec) & - & - & - & - & - & 0.53 / 0.59 &  - \\
		& Onset Latency & - & - & - & - & - & - & 15.83 \\

		\midrule
		\multirow{6}{*}{CNN1D + LSTM} 
		& OVLP & - & - & 1.0 & 0.52 & 72.13 & - & - \\
		& TAES & - & - & 0.21 & 0.99 & 6.6 & - & - \\
		& EPOCH & 0.8 & 0.79 & 0.69 & 0.75 & - & - & -  \\
		& MARGIN(3sec)  & - & - & - & - & - & 0.24 / 0.35 & -  \\
		& MARGIN(5sec) & - & - & - & - & - & 0.33 / 0.4 & -  \\
		& Onset Latency & - & - & - & - & - & - & 42.84 \\
		
		\midrule
		\multirow{6}{*}{CNN1D + BLSTM} 
		& OVLP & - & - & 1.0 & 0.48 & 82.44 & - & - \\
		& TAES & - & - & 0.22 & 0.99 & 4.71 & - & - \\
		& EPOCH & 0.8 & 0.78 & 0.73 & 0.71 & - & - & -  \\
		& MARGIN(3sec)  & - & - & - & - & - & 0.33 / 0.42 & -  \\
		& MARGIN(5sec) & - & - & - & - & - & 0.56 / 0.51 & - \\
		& Onset Latency & - & - & - & - & - & - & 43.64 \\
		\midrule
		\multirow{6}{*}{ResNet18} 
		& OVLP & - & - & 1.0 & 0.51 & 76.15 & - & - \\
		& TAES & - & - & 0.18 & 0.99 & 7.47 & - & - \\
		& EPOCH & 0.81 & 0.78 & 0.75 & 0.75 & - & - & -  \\
		& MARGIN(3sec)  & - & - & - & - & - & 0.24 / 0.28 & -  \\
		& MARGIN(5sec) & - & - & - & - & - & 0.30 / 0.34 & \\
		& Onset Latency & - & - & - & - & - & - & 20.99 \\
		\midrule
		\multirow{6}{*}{MobileNetV3} 
		& OVLP & - & - & 0.86 & 0.43 & 739.24 & - & - \\
		& TAES & - & - & 0.23 0.01 & 1.0 0.0 & 2.1 & - & - \\
		& EPOCH & 0.85 & 0.83 & 0.77 0.04 & 0.76 & - & - & -  \\
		& MARGIN(3sec)  & - & - & - & - & - & 0.38 / 0.46 & - \\
		& MARGIN(5sec) & - & - & - & - & - & 0.47 / 0.55 & - \\
		& Onset Latency & - & - & - & - & - & - & 15.66 \\
		\midrule
		\multirow{6}{*}{AlexNet} 
		& OVLP & - & - & 0.86 0.2 & 0.62 & 93.54 & - & - \\
		& TAES & - & - & 0.26 & 1.0 & 2.09 & - & - \\
		& EPOCH & 0.83 & 0.81 & 0.70 & 0.81 & - & - & -  \\
		& MARGIN(3sec)  & - & - & - & - & - & 0.46 / 0.55 & - \\
		& MARGIN(5sec) & - & - & - & - & - & 0.56 / 0.64 & - \\
		& Onset Latency & - & - & - & - & - & - & 12.07 \\
		\midrule
		\multirow{6}{*}{DenseNet} 
		& OVLP & - & - & 1.0 & 0.49 & 85.19  & - & - \\
		& TAES & - & - & 0.19 & 0.99 & 4.51 & - & - \\
		& EPOCH & 0.82& 0.79 & 0.78 0.04 & 0.73 & - & - & -  \\
		& MARGIN(3sec)  & - & - & - & - & - & 0.29 / 0.36 & - \\
		& MARGIN(5sec) & - & - & - & - & - & 0.37 / 0.45 & - \\
		& Onset Latency & - & - & - & - & - & - & 18.74 \\
		\midrule
		\multirow{6}{*}{ChronoNet} 
		& OVLP & - & - & 1.0 & 0.52 & 74.34 & - & - \\
		& TAES & - & - & 0.13 & 0.93 & 85.50  & - & - \\
		& EPOCH & 0.59 & 0.57 & 0.58 & 0.56 & - & - & -  \\
		& MARGIN(3sec)  & - & - & - & - & - & 0.20 / 0.25& -  \\
		& MARGIN(5sec) & - & - & - & - & - & 0.25 / 0.34 & -  \\
		& Onset Latency & - & - & - & - & - & - & 63.78 \\
		\midrule
		\multirow{6}{*}{TDNN+LSTM} 
		& OVLP & - & - & 1.0 & 0.52 & 80.55 & - & - \\
		& TAES & - & - & 0.20 & 0.98 & 18.66 & - & - \\
		& EPOCH & 0.80  & 0.78 & 0.72 & 0.73 & - & - & -  \\
		& MARGIN(3sec)  & - & - & - & - & - & 0.28 / 0.33  & -  \\
		& MARGIN(5sec) & - & - & - & - & - & 0.36 / 0.41 & -   \\
		& Onset Latency & - & - & - & - & - & - & 46.72 \\
		\midrule
		\multirow{6}{*}{Feature Transformer} 
		& OVLP & - & - & 1.0 & 0.53 & 73.22 & - & - \\
		& TAES & - & - & 0.16 & 0.89 & 110.96 & - & - \\
		& EPOCH & 0.6 & 0.6 & 0.46 & 0.72 & - & - & -  \\
		& MARGIN(3sec)  & - & - & - & - & - & 0.05 / 0.11 & - \\
		& MARGIN(5sec) & - & - & - & - & - & 0.08 / 0.13 & - \\
		& Onset Latency & - & - & - & - & - & - & 63.78 \\
		\midrule
		\multirow{6}{*}{Guided Feature Transformer} 
		& OVLP & - & - & 1.0 & 0.52 & 72.13 & - & - \\
		& TAES & - & - & 0.28 & 0.99 & 2.89 & - & - \\
		& EPOCH & 0.82 & 0.81 & 0.7 & 0.8 & - & - & -  \\
		& MARGIN(3sec)  & - & - & - & - & - & 0.43 / 0.48 & - \\
		& MARGIN(5sec) & - & - & - & - & - & 0.49 / 0.55 & -  \\
		& Onset Latency & - & - & - & - & - & - & 19.95 \\
		\bottomrule
	\end{tabular}
\end{table*}
We present seizure detector evaluation metric with multiple models (Table \ref{eval_models}, \ref{eval_preproc_appendix}, \ref{eval_preproc_appendix_resnetlstm}), which can further be used as the reference value for future studies.
\begin{table*}[ht!]
	\footnotesize
	\centering
	\caption{Exploration on four evaluation methods on real-time seizure detection trained with each different signal process feature extractor on CNN2D + LSTM. We evaluate TPR, TNR and FAs/24hours that maximizes TPR + TNR, and measured MARGIN and Latency when TNR is above 0.95.}
	\label{eval_preproc_appendix}
	\begin{tabular}{c|c|ccccccc}
		\toprule
		Feature Extraction & Metrics & AUROC & AUPRC & TPR & TNR & FAs / 24 hrs & Acc(Onset, Offset) & Time(Sec)\\
		\midrule
		\multirow{6}{*}{Raw} 
		& OVLP & - & - & 0.75 & 0.82 & 47.06 & - & - \\
		& TAES & - & - & 0.33 & 1.0 & 1.03 & - & - \\
		& EPOCH & 0.89 & 0.88 & 0.81 & 0.83 & - & - & - \\
		& MARGIN(3sec)  & - & - & - & - & - & 0.41, 0.5 & - \\
		& MARGIN(5sec) & - & - & - & - & - & 0.56, 0.51 & - \\
		& Onset Latency & - & - & - & - & - & - & 10.55 \\
		\midrule
		\multirow{6}{*}{Sincnet}
		& OVLP & - & - & 1.0 & 0.52 & 79.33 & - & - \\
		& TAES & - & - & 0.2 & 0.99 & 3.83 & - & - \\
		& EPOCH & 0.83 & 0.81 & 0.72 & 0.78 & - & - & - \\
		& MARGIN(3sec)  & - & - & - & - & - & 0.42, 0.45 & - \\
		& MARGIN(5sec) & - & - & - & - & - & 0.5, 0.52 & - \\
		& Onset Latency & - & - & - & - & - & - & 15.46 \\
		\midrule
		\multirow{6}{*}{STFT} 
		& OVLP & - & - & 0.67 & 0.9 & 34.13 & - & - \\
		& TAES & - & - & 0.29 & 0.99 & 4.3 & - & - \\
		& EPOCH & 0.91 & 0.90 & 0.85 & 0.82 & - & - & - \\
		& MARGIN(3sec)  & - & - & - & - & - & 0.39, 0.52 & - \\
		& MARGIN(5sec) & - & - & - & - & - & 0.55, 0.59 & - \\
		& Onset Latency & - & - & - & - & - & - & 2.39 \\
		\midrule
		\multirow{6}{*}{Frequency Bands} 
		& OVLP & - & - & 0.7 & 0.88 & 41.28 & - & - \\
		& TAES & - & - & 0.3 & 1.0 & 0.87 & - & - \\
		& EPOCH & 0.92 & 0.91 & 0.85 & 0.83 & - & - & - \\
		& MARGIN(3sec)  & - & - & - & - & - & 0.44, 0.48 & - \\
		& MARGIN(5sec) & - & - & - & - & - & 0.56, 0.57 & - \\
		& Onset Latency & - & - & - & - & - & - & 8.35 \\
		\midrule
		\multirow{6}{*}{Downsampled} 
		& OVLP & - & - & 0.63 & 0.87 & 42.94 & - & - \\
		& TAES & - & - & 0.26 & 0.99 & 4.81 & - & - \\
		& EPOCH & 0.88 & 0.87 & 0.78 & 0.83 & - & - & - \\
		& MARGIN(3sec)  & - & - & - & - & - & 0.45, 0.45 & - \\
		& MARGIN(5sec) & - & - & - & - & - & 0.53, 0.55 & - \\
		& Onset Latency & - & - & - & - & - & - & 11.52 \\
		\midrule
		\multirow{6}{*}{LFCC} 
		& OVLP & - & - & 0.66 & 0.85 & 62.75 & - & - \\
		& TAES & - & - & 0.26 & 0.99 & 4.81 & - & - \\
		& EPOCH & 0.86 & 0.84 & 0.73 & 0.84 & - & - & - \\
		& MARGIN(3sec)  & - & - & - & - & - & 0.41, 0.44 & - \\
		& MARGIN(5sec) & - & - & - & - & - & 0.52, 0.51 & - \\
		& Onset Latency & - & - & - & - & - & - & 11.52 \\
		\bottomrule
	\end{tabular}
\end{table*}

\begin{table}[h!]
    \small
	\centering
	\caption{Real-time seizure detection with different \textbf{window size} on bipolar TUH EEG dataset trained with Resnet-short+LSTM, averaged over $5$ runs.The CPU speed setting and meaning are explained in \ref{sec: speed}}
	\label{windowsize_resnetlstm}
	\begin{tabular}{c|ccc}
		\toprule
		Window & AUROC & AUPRC & Process\\
		(sec) &&& (sec)\\
		\midrule
		1 & 0.89 $\pm$ 0.01 & 0.88 $\pm$ 0.01 & \textbf{0.2423} \\
		2  & 0.9 $\pm$ 0.01 & 0.86 $\pm$ 0.06 & \textbf{0.5178} \\
		\textbf{4} & 0.92 $\pm$ 0.00 & 0.91 $\pm$ 0.00 & \textbf{0.941}\\
		6  & 0.9 $\pm$ 0.01 & 0.9 $\pm$ 0.01 & 1.2196 \\
		7  & 0.91 $\pm$ 0.01 & 0.9 $\pm$ 0.01 & 1.3012 \\
		8  & 0.9 $\pm$ 0.02 & 0.9 $\pm$ 0.00 & 1.3785 \\
		10  & 0.93 $\pm$ 0.01 & 0.92 $\pm$ 0.01 & 1.9242 \\
		12  & 0.92 $\pm$ 0.01 & 0.91 $\pm$ 0.01  & 2.0114 \\
		\bottomrule
	\end{tabular}
\end{table}
\begin{table}[h!]
	\small
	\centering
	\caption{Real-time seizure detection with different \textbf{window shift length} on bipolar TUH EEG dataset trained with Resnet-short+LSTM with fixed 4 second sliding window, averaged over $5$ runs.}
	\label{shift_resnetlstm}
	\begin{tabular}{c|ccc}
		\toprule
		Shift Length & AUROC & AUPRC \\
		(sec)\\
		\midrule
		1 & \textbf{0.92 $\pm$ 0.00} & \textbf{0.91 $\pm$ 0.00}\\
		2 & 0.9 $\pm$ 0.02 & 0.89 $\pm$ 0.02\\ 
		3 & 0.9 $\pm$ 0.01 & 0.89 $\pm$ 0.01\\
		4 & 0.89 $\pm$ 0.02 & 0.88 $\pm$ 0.02\\
		5 & 0.9 $\pm$ 0.01 & 0.89 $\pm$ 0.01\\
		\bottomrule
	\end{tabular}
\end{table}

\begin{table*}[ht!]
	\footnotesize
	\centering
	\caption {Exploration on four evaluation methods on real-time seizure detection trained with each different signal process feature extractor on Resnet-Short + LSTM. We evaluate TPR, TNR and FAs/24hours that maximizes TPR + TNR, and measured MARGIN and Latency when TNR is above 0.95.}
	\label{eval_preproc_appendix_resnetlstm}
	\begin{tabular}{c|c|ccccccc}
		\toprule
		Feature Extraction & Metrics & AUROC & AUPRC & TPR & TNR & FAs / 24 hrs & Acc(Onset, Offset) & Time(Sec)\\
		\midrule
		\multirow{6}{*}{Raw} 
		& OVLP & - & - & 0.75 & 0.84 & 64.9 & - & - \\
		& TAES & - & - & 0.4 & 1.0 & 1.68 & - & - \\
		& EPOCH & 0.92 & 0.91 & 0.83 & 0.85 & - & - & -  \\
		& MARGIN(3sec)  & - & - & - & - & - & 0.49 / 0.58 & -  \\
		& MARGIN(5sec) & - & - & - & - & - & 0.62 / 0.65 & -  \\
		& Onset Latency & - & - & - & - & - & - & 15.23 \\
		\midrule
		\multirow{6}{*}{Sincnet}
		& OVLP & - & - & 1.0 & 0.52 & 81.14 & - & - \\
		& TAES & - & - & 0.33 & 0.99 & 4.87 & - & - \\
		& EPOCH & 0.86 & 0.76 & 0.74 & 0.81 & - & - & - \\
		& MARGIN(3sec)  & - & - & - & - & - & 0.38, 0.45 & - \\
		& MARGIN(5sec) & - & - & - & - & - & 0.51, 0.57 & - \\
		& Onset Latency & - & - & - & - & - & - & 24.77 \\
		\midrule
		\multirow{6}{*}{STFT} 
		& OVLP & - & - & 1.0 & 0.52 & 74.3 & - & - \\
		& TAES & - & - & 0.34 & 0.99 & 6.12 & - & - \\
		& EPOCH & 0.91 & 0.8 & 0.85 & 0.83 & - & - & - \\
		& MARGIN(3sec)  & - & - & - & - & - & 0.35, 0.48 & - \\
		& MARGIN(5sec) & - & - & - & - & - & 0.45, 0.52 & - \\
		& Onset Latency & - & - & - & - & - & - & 8.19 \\
		\midrule
		\multirow{6}{*}{Frequency Bands} 
		& OVLP & - & - & 0.85 & 0.713 & 161.6 & - & - \\
		& TAES & - & - & 0.37 & 1.0 & 1.39 & - & - \\
		& EPOCH & 0.92 & 0.86 & 0.87 & 0.87 & - & - & - \\
		& MARGIN(3sec)  & - & - & - & - & - & 0.46, 0.56 & - \\
		& MARGIN(5sec) & - & - & - & - & - & 0.59, 0.63 & - \\
		& Onset Latency & - & - & - & - & - & - & 10.8 \\
		\midrule
		\multirow{6}{*}{Downsampled} 
		& OVLP & - & - & 0.74 & 0.83 & 81.57 & - & - \\
		& TAES & - & - & 0.41 & 1.0 & 2.65 & - & - \\
		& EPOCH & 0.92 & 0.87 & 0.85 & 0.87 & - & - & - \\
		& MARGIN(3sec)  & - & - & - & - & - & 0.38, 0.45 & - \\
		& MARGIN(5sec) & - & - & - & - & - & 0.51, 0.57 & - \\
		& Onset Latency & - & - & - & - & - & - & 3.96 \\
		\midrule
		\multirow{6}{*}{LFCC} 
		& OVLP & - & - & 0.75 & 0.67 & 184.68 & - & - \\
		& TAES & - & - & 0.3 & 0.99 & 4.74 & - & - \\
		& EPOCH & 0.92 & 0.84 & 0.8 & 0.88 & - & - & - \\
		& MARGIN(3sec)  & - & - & - & - & - & 0.43, 0.48 & - \\
		& MARGIN(5sec) & - & - & - & - & - & 0.52, 0.55 & - \\
		& Onset Latency & - & - & - & - & - & - & 13 \\
		\bottomrule
	\end{tabular}
\end{table*}

\begin{table*}[ht!]
	\small
	\centering
	\caption{Evaluation on anti-aliasing effect: 0 to 100Hz Band-pass filtering is applied before down-sampling feature extraction on \textit{raw} bipolar TUH EEG V1.5.2 dataset to see the effect of anti-aliasing on performance. The results were averaged over $5$ runs.}
	\label{antialiasing}
	\begin{tabular}{c|c|ccccc}
		\toprule
		Feature Extraction & Model Type & AUROC & AUPRC & TPR & TNR\\
		\midrule
		Downsampled feature extraction
		&CNN2D+LSTM 
		& 0.88 $\pm$ 0.04
		& 0.87 $\pm$ 0.04
		& 0.74 $\pm$ 0.06 
		& 0.87 $\pm$ 0.00
		\\
		after bandpass filter &ResNet-short + LSTM
		& 0.91 $\pm$ 0.00 
		& 0.9 $\pm$ 0.00
		& 0.83 $\pm$ 0.01 
		& 0.82 $\pm$ 0.02 \\
	
		\midrule
		\multirow{2}{*}{Downsampled feature extraction}
		&CNN2D+LSTM 
		& 0.88 $\pm$ 0.02
		& 0.87 $\pm$ 0.02
		& 0.78 $\pm$ 0.04
		& 0.83 $\pm$ 0.03
		\\
		&ResNet-short + LSTM
		& 0.92 $\pm$ 0.01 
		& 0.87 $\pm$ 0.02 
		& 0.85 $\pm$ 0.05 
		& 0.86 $\pm$ 0.04
		\\

		\bottomrule
	\end{tabular}
\end{table*}

\end{document}